\documentclass[journal]{IEEEtran}

\usepackage[T1]{fontenc}
\usepackage[utf8]{inputenc}
\usepackage{textcomp}
\usepackage{amsmath,amsfonts,amssymb,amsthm}
\usepackage{bm}
\usepackage{cite}
\usepackage{graphicx}
\usepackage{svg} 
\usepackage[caption=false,font=normalsize,labelfont=sf,textfont=sf]{subfig}
\usepackage{algorithm}
\usepackage{algorithmic}
\usepackage{array}
\usepackage{booktabs} 
\usepackage{stfloats} 
\usepackage{url}
\usepackage{verbatim}

\usepackage[table]{xcolor}
\definecolor{myred}{RGB}{255, 0, 0}
\definecolor{myblue}{RGB}{0, 0, 200}
\definecolor{lightblue}{RGB}{220, 235, 250}
\definecolor{lightpink}{RGB}{255, 235, 245}
\definecolor{lightflesh}{RGB}{255, 235, 220}
\newcommand{\best}[1]{\textcolor{myred}{\textbf{#1}}}
\newcommand{\second}[1]{\textcolor{myblue}{\textbf{#1}}}

\DeclareMathOperator{\argmin}{arg\,min}

\DeclareMathOperator{\Tr}{Tr}

\begin{document}

\title{Federated Multi-Task Clustering}

\author{Suyan~Dai, 
        Gan~Sun*,~\IEEEmembership{Member~IEEE}, 
        Fazeng~Li, 
        Xu~Tang, 
        Qianqian~Wang, 
        and~Yang~Cong~\IEEEmembership{Member,~IEEE}%

\thanks{Suyan Dai, Gan Sun, Fazeng Li and Yang Cong are with the School of Automation Science and Engineering, South China University of Technology, Guangzhou 510641, China. (email: 18790457702@163.com, sungan1412@gmail.com, lifazeng818@gmail.com, congyang81@gmail.com)}

\thanks{Xu Tang is with the School of Software, Dalian University of Technology, Dalian City 116024, China. (email: tangxu@mail.dlut.edu.cn)}
\thanks{Qianqian Wang is with the School of Telecommunications Engineering, Xidian University, Xi'an, 710071, China.(email: qqwang@xidian.edu.cn)}

\thanks{This work is supported by the National Key Research and Development Program of China (2022YFC2806101) and National Nature Science Foundation of China under Grant (62273333, 62225310, 62127807), and the Fundamental Research Funds for the Central Universities (2024ZYGXZR024).}
\thanks{$^{*}$The corresponding author is \emph{Prof. Gan Sun}.}
}

\markboth{Journal of \LaTeX\ Class Files,~Vol.~14, No.~8, August~2021}%
{Shell \MakeLowercase{\textit{et al.}}: A Sample Article Using IEEEtran.cls for IEEE Journals}

\IEEEpubid{0000--0000/00\$00.00~\copyright~2021 IEEE}

\maketitle

\begin{abstract}
Spectral clustering has emerged as one of the most effective clustering algorithms due to its superior performance. However, most existing models are designed for centralized settings, rendering them inapplicable in modern decentralized environments. Moreover, current federated learning approaches often suffer from poor generalization performance due to reliance on unreliable pseudo-labels, and fail to capture the latent correlations amongst heterogeneous clients. To tackle these limitations, this paper proposes a novel framework named \underline{F}ederated \underline{M}ulti-\underline{T}ask \underline{C}lustering (\emph{i.e.,} \textbf{FMTC}), which intends to learn personalized clustering models for heterogeneous clients while collaboratively leveraging their shared underlying structure in a privacy-preserving manner. More specifically, the FMTC framework is composed of two main components: client-side personalized clustering module, which learns a parameterized mapping model to support robust out-of-sample inference, bypassing the need for unreliable pseudo-labels; and server-side tensorial correlation module, which explicitly captures the shared knowledge across all clients. This is achieved by organizing all client models into a unified tensor and applying a low-rank regularization to discover their common subspace. To solve this joint optimization problem, we derive an efficient, privacy-preserving distributed algorithm based on the Alternating Direction Method of Multipliers, which decomposes the global problem into parallel local updates on clients and an aggregation step on the server. To the end, several extensive experiments on multiple real-world datasets demonstrate that our proposed FMTC framework significantly outperforms various baseline and state-of-the-art federated clustering algorithms.
\end{abstract}


%%%%%%%%%%%%%%%%%%%%%%%%%%%%%%%%%%%%%%%%%%%%%%%%%%%%%%%%%%%%%%%%%%%%%%%%%%%%%%%%%%
\section{Introduction}
\IEEEPARstart{S}{pectral} clustering \cite{ng2001spectral} is a powerful and fundamental algorithm that has demonstrated state-of-the-art performance in numerous applications. It operates by effectively discovering high-quality embeddings from the manifold structure inherent in the data distribution. Recent research has extended the boundaries of spectral clustering to more complex scenarios. For example, researchers have proposed various scalable variants \cite{zhang2025anchor, feng2025incremental} to address computational challenges. To learn more discriminative feature representations from raw data, deep spectral clustering improves clustering performance through the organic integration of deep neural networks and spectral analysis \cite{zhao2025deep, williams2025spectral}. However, most existing spectral clustering algorithms rely on a critical assumption: data can be freely collected and processed by a central server. This centralized requirement renders them inapplicable when data is distributed across different institutions or user devices due to privacy constraints. Consequently, how to adapt the principles of federated learning to clustering methods has emerged as a novel paradigm.

\IEEEpubidadjcol

\begin{figure}[!t]
    \centering
    \includegraphics[width=\columnwidth]{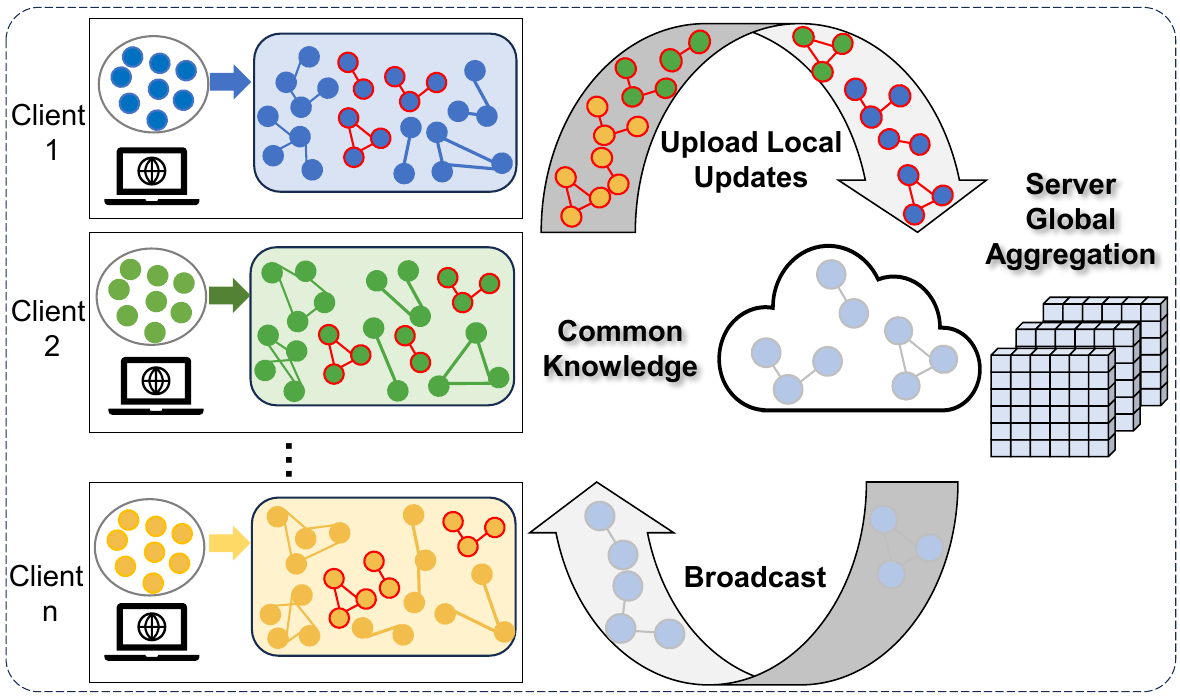}
    \caption{Motivation of our federated multi-task clustering model, where clients in each round share local model updates and the server aggregates these into a model tensor with multi-task constraint, distilling shared structure which is then broadcast back to refine each personalized model.}
    \label{fig:conceptual}
\end{figure}

Federated learning (FL) \cite{mcmahan2017communication} has emerged as a prominent distributed paradigm, enabling collaborative model training while preserving data privacy. In the context of unsupervised federated clustering, most existing works \cite{han2022fedx, servetnyk2020unsupervised} have largely adopted strategies from the broader FL field. These strategies typically involve either aggregating local clustering results into a unified global model \cite{kumar2020federated, garst2024federated, qiao2023federated}, or adopting personalized approaches to learn distinct models for different clients \cite{tan2022towards, deng2020adaptive}. However, these methods often fail to adequately account for the underlying relationships that exist among clients within the same clustering context. For example, in cross-institution web page clustering (\emph{e.g.,} the WebKB dataset), while the data distribution of each university (client) is distinct, requiring a personalized solution, they simultaneously share an underlying semantic structure (\emph{e.g.,} "Faculty page" and "Course page"). An advanced federated clustering method that can both learn personalized models and precisely capture this shared structure would significantly enhance performance across all tasks.

Inspired by the aforementioned observations, we in this paper propose a novel federated clustering learning framework, which takes two types of relations into consideration: 1) \textbf{Intra-client Correlation}, which refers the correlations amongst heterogeneous data in each client, and the process of learning cluster labels. In constrast, most existing federated clustering methods rely heavily on unreliable pseudo-labels for training and struggle with out-of-sample generalization; and 2) \textbf{Inter-client Correlation}, the latent common knowledge should be consistent among heterogeneous clients, which resolves the limitation that the majority of existing federated clustering methods. For example, in the 20NewsGroups dataset, each client owns articles from different thematic forums. These corpora follow non-overlapping vocabularies and discourse styles, so collapsing them into a single spectral model erases crucial client-specific cues. Yet they still share latent communicative patterns, which FMTC captures through its tensor regularizer while retaining personalized projections for every forum.

To tackle these dual challenges, as shown in Fig.~\ref{fig:conceptual}, we propose a \underline{F}ederated \underline{M}ulti-\underline{T}ask \underline{C}lustering (\emph{i.e.,} {FMTC}), which can achieve federated clustering knowledge transfer among heterogeneous clients. To achieve this goal, two components of our FMTC model are proposed to preserve the common information among multiple clients, \emph{i.e.,} client-side personalized clustering module and a server-side tensorial correlation module. To be specific, the personalized clustering module in each client integrates the spectral embedding learning and the mapping function optimization into a unified closed-loop system, which can capture the intrinsic manifold structure of the local client data, and be able to out-of-sample data. Based on the assumption that multiple client models are lie within a shared global subspace, the server-side tensorial correlation module mathematically organizes the local model parameters from all clients into a third-order tensor, and apply a low-rank regularization to constrain inter-client correlation. This module could effectively filter out client-specific noise while enforcing a global consensus on the model structure, thereby allowing the shared knowledge to reinforce the performance of each individual client. To solve this optimization problem, we derive an efficient and privacy-preserving distributed algorithm based on the Alternating Direction Method of Multipliers (ADMM). This algorithm decomposes the complex global objective into parallel local updates on clients and a global aggregation step on the server, ensuring convergence without exchanging raw data. To the end, our FMTC model is evaluated against several spectral clustering algorithms and federated clutering learning models on several datasets. The experimental results strongly support our proposed FMTC model.

Our main contributions are summarized as follows:
\begin{itemize}
\item We propose a \underline{F}ederated \underline{M}ulti-\underline{T}ask \underline{C}lustering (\emph{i.e.,} {FMTC}) framework, which uniquely bridges the gap between personalized local adaptation and global collaboration, effectively resolving the dual challenges of unreliable pseudo-label dependency within clients and the neglect of structural correlations among clients.
\item We design a novel {bi-level collaborative strategy} that seamlessly integrates the {client-side personalized clustering module with the server-side tensorial correlation module}.This design allows for the direct learning of robust mapping functions for out-of-sample inference while explicitly capturing the high-order shared knowledge across heterogeneous clients via low-rank tensor regularization.
\item We develop a privacy-preserving distributed optimization algorithm based on ADMM to solve the distinct nonconvex joint problem efficiently. Extensive experiments on seven benchmarks validate that FMTC outperforms state-of-the-art methods, particularly in scenarios requiring strong generalization capabilities on unseen data.
\end{itemize}

The remainder of this paper is organized as follows. Section~\ref{sec:related_work} reviews the related literature on spectral clustering and federated clustering. Section~\ref{sec:method} details the proposed FMTC framework, including the mathematical formulation of the local and global objectives, followed by the the distributed ADMM optimization algorithm in Section~\ref{sec:optimization}. Section~\ref{sec:experiments} presents the experimental results and Section~\ref{sec:conclusion} concludes the paper and discusses future research directions.

%%%%%%%%%%%%%%%%%%%%%%%%%%%%%%%%%%%%%%%%%%%%%%%%%%%%%%%%%%%%
\section{Related Work}\label{sec:related_work}
In this section, we review the literature relevant to our work with two main streams: Spectral Clustering and Federated Clustering, and discuss the recent advancements in these fields.
%and identify the specific limitations of existing approaches—particularly regarding their inability to simultaneously handle data heterogeneity and task relatedness—which motivate the design of our proposed FMTC framework.

\subsection{Spectral Clustering}
Spectral clustering is a powerful graph clustering technique, widely applied for its ability to effectively identify non-convex clusters and reveal the intrinsic manifold structure of the data \cite{ng2001spectral}. Its success is particularly notable in various multimedia applications, such as image segmentation, video object clustering, and community detection in social networks. As the field matured, research has focused on advancing its core components. One line of work has concentrated on the construction of similarity graphs, leading to the development of adaptive neighbor selection techniques and local density-sensitive similarity measures to create more robust graph representations \cite{xie2024consistent, zhang2025anchor}. Another direction explores variants of the Laplacian matrix itself to better capture the local manifold structure of the data \cite{garcia2025eplsc, zhang2024enhancing}. To improve applicability for large-scale data, significant efforts have been made in developing approximation methods to improve computational efficiency \cite{feng2025incremental, rouhi2024two}. Furthermore, the integration of deep learning has led to deep spectral clustering, which uses neural networks to learn an optimal embedding space prior to spectral analysis \cite{guo2024deep, zhao2025deep}. In parallel, deep multi-view clustering methods have explored learning discriminative representations to enhance inter-cluster separability while maintaining cross-view consistency\cite{10884978}. Despite this rich body of work advances its different components, a common characteristic of these models is their centralized design. The execution of the algorithm requires unrestricted access to the entire dataset. This property poses a significant challenge for applications in distributed scenarios with strict privacy and data sovereignty requirements.

\subsection{Federated Clustering Learning}
Research on clustering in federated learning \cite{mcmahan2017communication} can be broadly divided into two categories. The first category uses clustering as an auxiliary tool for handling non-independent and identically distributed (Non-IID) data, with the primary goal of improving the performance of downstream supervised learning tasks \cite{presotto2023federated, li2021federated}. In these approaches, the system typically groups clients (or devices) to train more adaptive supervised models for each group \cite{ghosh2022efficient, sattler2020clustered, zhang2025lcfed}. The second category focuses directly on the unsupervised clustering task itself \cite{Zhang2024AsynchronousFC}. This line of research aims to find meaningful cluster structures from distributed data while preserving privacy. Such methods hold great promise for multimedia scenarios, for instance, in building personalized media recommendation systems or analyzing user behavior across different platforms without sharing raw data. 

Common technical approaches include the iterative aggregation of local models or parameters, such as in Federated K-Means \cite{kumar2020federated, garst2024federated}, and methods based on distributed matrix factorization \cite{yfantis2025federated, feng2024efficient}. Additionally, adaptive weighting strategies have proven effective in multi-view scenarios for distinguishing the importance of heterogeneous data sources and fusing complementary information\cite{9312643}. Beyond model design and aggregation strategies, the robustness of federated learning systems under adversarial settings has also attracted increasing attention. For example, Sun \emph{et al.} \cite{9618642} showed that federated multi-task learning is vulnerable to data poisoning attacks, indicating that robustness and security remain important challenges in distributed multi-client learning. However, a common limitation persists across both categories of existing federated clustering methods. They typically stop at learning either a single global model or a set of independent personalized models, failing to explicitly model the underlying deep relationships between clients as a core component of the framework. To the best of our knowledge, there are no existing works that simultaneously learn personalized models and explicitly capture global task relationships within a unified federated clustering setting. Our work aims to fill this critical gap by introducing a multi-task learning paradigm, representing the first effort to achieve federated multi-task clustering.

\section{Our Proposed Method}\label{sec:method}

\begin{figure*}[!t]
    \centering
    \includegraphics[width=\textwidth]{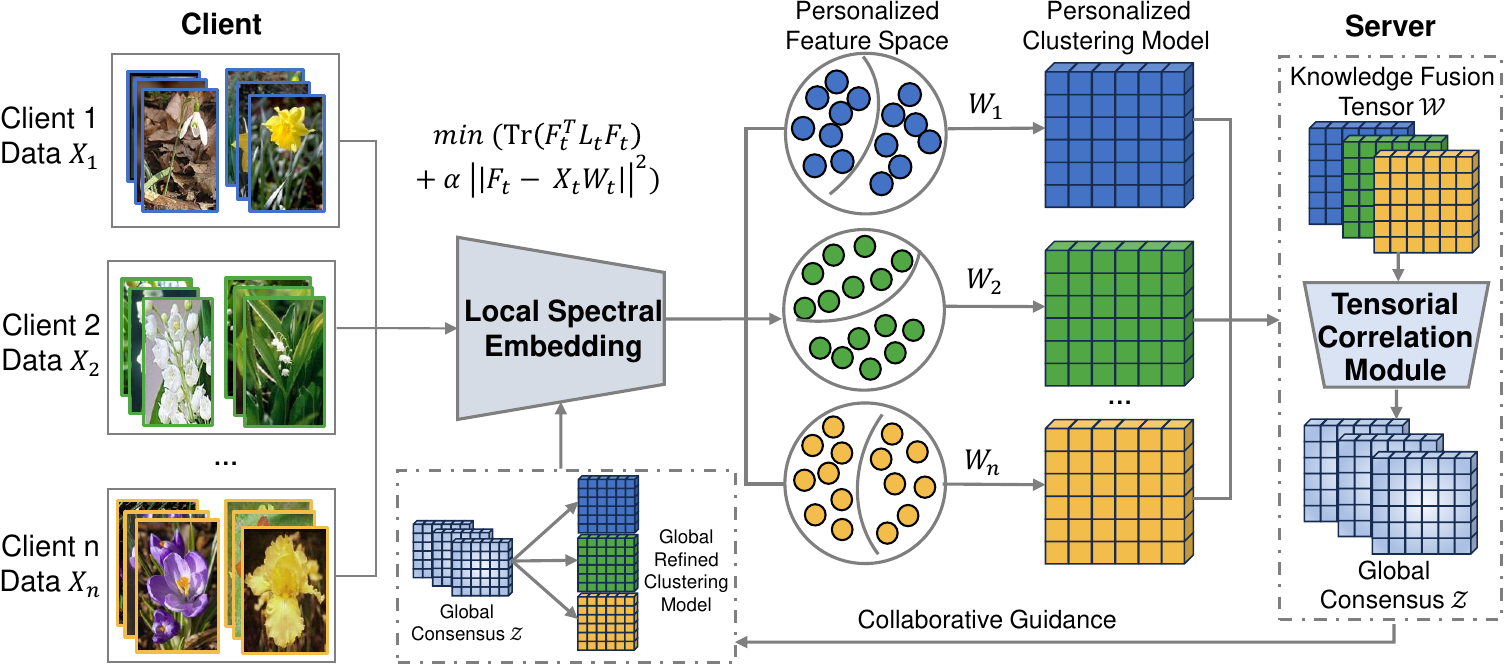}
    \caption{The architecture of our Federated Multi-Task Clustering (FMTC) framework, where the process begins with heterogeneous data samples $\{X_t\}$ from diverse clients.
    At the client level, each client data is processed through Local Spectral Embedding, which is optimized according to our local objective function. This process generates a client-specific Personalized Feature Space and learns a corresponding Personalized Clustering Model $W_t$. At the server level, all personalized clustering models $\{W_t\}$ are collected and stacked into a Knowledge Fusion Tensor $\bm{\mathcal{W}}$. The Tensorial Correlation Module is then applied to this tensor to distill a Global Consensus $\bm{\mathcal{Z}}$, representing the shared knowledge across all tasks. This consensus is further used to obtain the Global Refined Clustering Model. Finally, the Collaborative Guidance from the server provides regularization back to the local spectral embedding, completing the federated optimization loop.}
    \label{fig:framework}
\end{figure*}

\subsection{Preliminaries}
Given a data set of $n$ points with dimensions $d$, $\mathbf{X} = [\mathbf{x}_1, \dots, \mathbf{x}_n] \in \mathbb{R}^{d \times n}$, the clustering task in each client aims to partition these points into $c$ groups. Spectral clustering \cite{ng2001spectral} constructs a similarity graph to encode pairwise relationships as a similarity graph. The structure of the graph is captured by a weighted adjacency matrix $\mathbf{A} \in \mathbb{R}^{n \times n}$, where $A_{ij} > 0$ quantifies the affinity between points $\mathbf{x}_i$ and $\mathbf{x}_j$. A common construction method is the graph of the $k$-nearest neighbor ($k$-NN), where edge weights are defined by a Gaussian kernel:
\begin{equation}
A_{ij} = 
\begin{cases}
    e^{-\frac{\|\mathbf{x}_i - \mathbf{x}_j\|^2}{2\sigma^2}} & \text{if } \mathbf{x}_i \in \mathcal{N}_k(\mathbf{x}_j) \lor \mathbf{x}_j \in \mathcal{N}_k(\mathbf{x}_i) \\
    0 & \text{otherwise},
\end{cases}
\end{equation}
and $\mathcal{N}_k(\cdot)$ is the set of k-nearest neighbors and $\sigma$ is the kernel bandwidth. We can compute the diagonal degree matrix $\mathbf{D}$ from $\mathbf{A}$, where $D_{ii} = \sum_j A_{ij}$. The unnormalized Laplacian graph is then defined as $\mathbf{L} = \mathbf{D} - \mathbf{A}$. For improved numerical stability and its connection to the Normalized Cut objective, we use the symmetric normalized Laplacian $\hat{\mathbf{L}}$:
\begin{equation}
    \hat{\mathbf{L}} = \mathbf{D}^{-1/2} \mathbf{L} \mathbf{D}^{-1/2} = \mathbf{I} - \mathbf{D}^{-1/2} \mathbf{A} \mathbf{D}^{-1/2}.
\end{equation}

The final objective of spectral clustering is to find a discrete partition of the data. This can be formally expressed as an optimization problem to find an indicator matrix $\mathbf{Y} \in \mathbb{R}^{n \times c}$, where $c$ is the number of clusters. The problem can be formulated as:
\begin{equation}
    \min_{\mathbf{Y}} \quad \Tr(\mathbf{Y}^{\top} \hat{\mathbf{L}} \mathbf{Y}) \quad \text{s.t.} \quad \mathbf{Y} \in \text{Idx},
\end{equation}
where the set $\text{Idx}$ is the set of all valid partition matrices.

%%\begin{equation}
 %%   \text{Idx} = \left\{ \mathbf{Y} \in \{0, 1\}^{n \times c} \;\middle|\; \sum_{j=1}^c y_{ij} = 1, \forall i \in \{1, \dots, n\} \right\}.
%% \end{equation}
%%This constraint enforces that each row of $\mathbf{Y}$ is a \textbf{one-hot vector}, assigning each data point to exactly one of the $c$ clusters.

\textbf{Problem Definition:} Our work considers a federated learning scenario consisting of a central server $S$ and $m$ clients $\{C_t\}_{t=1}^m$. The data is distributed across these clients, where each client $C_t$ holds a local dataset $\mathbf{X}_t \in \mathbb{R}^{n_t \times d_t}$. Here, $n_t$ is the number of local samples and $d_t$ is the feature dimensionality for client $t$. We formulate the clients as distinct yet correlated tasks within a multi-task Learning paradigm~\cite{9392366}. The server $S$ is responsible for coordinating the clients to collaboratively learn their respective clustering structures without direct access to their local data. Our method aims to learn a distinct and high-quality clustering for each client from the decentralized local data, while leveraging shared knowledge across clients in a privacy-preserving manner. We consider a setting where that all the clients share a common feature space although the number of samples $n_t$ can vary across clients, \emph{i.e.,} the feature dimensionality $d_t = d$ for all clients t. This is a common setting in cross-silo federated learning~\cite{huang2022cross, huang2021personalized}.

%%%%%%%%%%%%%%%%%%%%%%%%%%%%%%%%%%%%%%%%%%%%%%%%%%%%%%%%%%%
\subsection{Federated Multi-Task Clustering}
Our Federated Multi-Task Clustering (FMTC) framework, whose overall architecture is depicted in Fig.~\ref{fig:framework}, consists of two core components: local model training conducted by each client and a global objective to ensure consistency and knowledge sharing, coordinated by the server.
%%%%%%%%%%%%%%%%%%%%%%%%%%%%%%%%%%%%%%%%%%%%%%%%%%%
\subsubsection{Client-Side Personalized Clustering Module}
For the client-side personalized clustering module, the primary objective is to perform clustering on its local data $\mathbf{X}_t$. Existing federated solutions\cite{Wen_CIKM_2021, Xu_AAAI_2022} often resort to generating static pseudo-labels to train a secondary classifier, but this ``two-stage'' process suffers from severe error propagation. Recent multi-view clustering studies have also recognized this limitation and proposed methods to directly derive discriminative clustering indicator matrices without additional post-processing\cite{10076472}. Moreover, traditional spectral clustering is non-parametric, yielding only embeddings $\mathbf{F}_t$ defined on specific training points, which makes out-of-sample inference impossible. To address the intra-client challenge regarding the reliance on unreliable pseudo-labels and poor generalization performance, we redesign the local learning objective in each client.

Our client-side personalized clustering module proposes to learn a parametric projection matrix $\mathbf{W}_t \in \mathbb{R}^{d_t \times k}$ simultaneously with the spectral embeddings. We posit that the ideal embedding $\mathbf{F}_t$ can be approximated by a linear transformation of the data $\mathbf{X}_t$. By enforcing the constraint $\| \mathbf{F}_t - \mathbf{X}_t \mathbf{W}_t \|_F^2$, we couple the manifold discovery and the mapping function learning into a unified optimization process. This design ensures that $\mathbf{W}_t$ directly captures the intrinsic data structure without needing noisy pseudo-labels, enabling the model to robustly generalize to unseen data. The unified local objective is formulated as:
\begin{equation}
\begin{aligned}
\min_{\mathbf{F}_t, \mathbf{W}_t} & \quad \Tr(\mathbf{F}_t^{\top} \hat{\mathbf{L}}_t \mathbf{F}_t) + \alpha \| \mathbf{F}_t - \mathbf{X}_t \mathbf{W}_t \|_F^2 \\ 
\text{s.t.} &\quad\mathbf{F}_t^{\top} \mathbf{F}_t = \mathbf{I},
\label{eq:local_objective}
\end{aligned}
\end{equation}
where the hyperparameter $\alpha$ controls the trade-off between preserving the ideal spectral structure (the first term) and ensuring that this structure is faithfully captured by a linear model.

%%%%%%%%%%%%%%%%%%%%%%%%%%%%%%%%%%%%%%%%%%%%%%%%%%%%
\subsubsection{Server-Side Tensorial Correlation Module}
The server-side tensorial correlation module of our framework is designed to tackle the {inter-client challenge}, where existing methods fail to leverage the latent correlations among heterogeneous clients. We hypothesize that while local models $\{\mathbf{W}_t\}$ are personalized to distinct data distributions, they share a common underlying structure governed by the semantic nature of the tasks\cite{8587123, WEN2021207}. This assumption aligns with recent findings in subspace clustering, where low-rank representation learning has been shown to effectively capture shared subspace structures and enhance intra-cluster cohesion\cite{10552250}.

%To explicitly model this relationship, the server-side tensorial correlation module organizes the collection of local models into a third-order tensor $\bm{\mathcal{W}} \in \mathbb{R}^{d \times k \times m}$. Unlike simple parameter averaging, we impose a low-rank constraint on $\bm{\mathcal{W}}$ to capture the high-order correlations across clients. 

To explicitly model this relationship, the server-side tensorial correlation module organizes the collection of local models into a third-order tensor $\bm{\mathcal{W}} \in \mathbb{R}^{d \times k \times m}$. This tensorization strategy is inspired by recent advances in ensemble clustering, where stacking multiple matrices into a tensor and minimizing the tensor Schatten p-norm has proven effective in extracting common shared information while eliminating noise\cite{10745551}. Unlike simple parameter averaging, we impose a low-rank constraint on $\bm{\mathcal{W}}$ to capture the high-order correlations across clients. Inspired by recent advances demonstrating the effectiveness of tensor Schatten p-norm in capturing complementary information and structural correlations across multiple views\cite{10858065}, we adopt this regularization to distill the shared knowledge among heterogeneous clients.This tensor-based regularization\cite{Wen_AAAI_2021} acts as a global knowledge sharing mechanism: it filters out client-specific noise and distills the shared structural knowledge into a common low-dimensional subspace. This ensures that the learning of each client is supported by the collective wisdom of all tasks. The global regularization term is defined as:
\begin{equation}
\label{eq:global_regularizer}
\begin{aligned}
    \min_{ \{ \mathbf{W}_t \}_{t=1}^m, \bm{\mathcal{W}} } & \quad \beta \| \bm{\mathcal{W}} \|_{S_p}^p \\
    \text{s.t.} & \quad \bm{\mathcal{W}}_{:,:,t} = \mathbf{W}_t, \quad \forall t \in \{1, \dots, m\}
\end{aligned}
\end{equation}
where $\beta$ is a hyperparameter that controls the strength of the collaboration. A larger $\beta$ forces the client models to be more similar, while a smaller $\beta$ allows for greater personalization. This term, managed by the server with access to all $\{\mathbf{W}_t\}$, forms the core of our collaborative learning strategy.
\subsubsection{Overall Objective Function}
By combining the local client objectives (Eq.~\eqref{eq:local_objective}) and the global collaboration regularizer (Eq.~\eqref{eq:global_regularizer}), we can formulate the complete optimization problem for our FMTC framework as follows:
\begin{equation}
\label{eq:full_objective_initial}
\begin{aligned}
\min_{\{\mathbf{F}_t\}, \{\mathbf{W}_t\}} \quad & \sum_{t=1}^{m} \omega_t \left( \Tr(\mathbf{F}_t^{\top} \hat{\mathbf{L}}_t \mathbf{F}_t) + \alpha \| \mathbf{F}_t - \mathbf{X}_t \mathbf{W}_t \|_F^2 \right) \\
& + \beta \|\bm{\mathcal{W}}\|_{S_p}^p \\
\text{s.t.} &\quad \mathbf{F}_t^{\top} \mathbf{F}_t = \mathbf{I},
\end{aligned}
\end{equation}
where $\bm{\mathcal{W}}$ is the tensor formed by stacking $\{\mathbf{W}_t\}$. This joint optimization problem is challenging to solve directly in a federated setting. The primary difficulty lies in the coupling introduced by the tensor norm regularizer $\beta \|\bm{\mathcal{W}}\|_{S_p}^p$, which depends on the parameters $\{\mathbf{W}_t\}$ of \textit{ all} clients simultaneously. This prevents clients from optimizing their local objectives in a fully parallel and independent manner.

\section{Model Optimization Framework}\label{sec:optimization}
To decouple the interdependent tensor constraints, we employ an alternate direction method of multipliers with variable splitting: we introduce a global auxiliary variable $\bm{\mathcal{Z}} \in \mathbb{R}^{d \times k \times m}$, which serves as a ``consensus'' copy of the model tensor $\bm{\mathcal{W}}$. We then enforce the constraint that the local models must agree with this global consensus, i.e. $\mathbf{W}_t = \mathbf{Z}_t$ for each client $t$, where $\mathbf{Z}_t$ is the $t$-th slice of $\bm{\mathcal{Z}}$. This allows us to reformulate Eq.~\eqref{eq:full_objective_initial} into an equivalent, constrained form that is separable:
\begin{equation}
\label{eq:admm_formulation}
\begin{aligned}
\min_{\{\mathbf{F}_t\}, \{\mathbf{W}_t\}, \bm{\mathcal{Z}}} & \sum_{t=1}^{m} \omega_t \left( \Tr(\mathbf{F}_t^{\top} \hat{\mathbf{L}}_t \mathbf{F}_t) + \alpha \| \mathbf{F}_t - \mathbf{X}_t \mathbf{W}_t \|_F^2 \right) \\
& + \beta \|\bm{\mathcal{Z}}\|_{S_p}^p \\
\text{s.t.} \quad & \mathbf{W}_t = \mathbf{Z}_t,\ \mathbf{F}_t^{\top} \mathbf{F}_t = \mathbf{I}, \quad \forall t \in \{1, \dots, m\}. \\
\end{aligned}
\end{equation}
Notice how the problematic coupling term now only involves the global variable $\bm{\mathcal{Z}}$, while the summation part only involves local variables $\{\mathbf{W}_t, \mathbf{F}_t\}$. This separable structure is ideal for the ADMM framework, which addresses the constraints by forming an Augmented Lagrangian. 
\subsection{Augmented Lagrangian}
To solve the constrained problem in Eq.~\eqref{eq:admm_formulation}, we construct its \textbf{Augmented Lagrangian} with two steps. For the set of equality constraints $\mathbf{W}_t = \mathbf{Z}_t$, we firstly introduce the corresponding set of dual variables $\{\mathbf{Y}_t\}_{t=1}^m$. For notational convenience, we can stack these matrices to form a single dual variable tensor, $\bm{\mathcal{Y}}$. Second, we add a quadratic penalty term to improve convergence properties. The resulting augmented Lagrangian $L_{\rho}$ is given by:
\begin{equation}
\label{eq:augmented_lagrangian}
\begin{aligned}
& L_{\rho}(\{\mathbf{F}_t,  \mathbf{W}_t\},  \bm{\mathcal{Z}}, \bm{\mathcal{Y}}) \\
& = \sum_{t=1}^{m} \omega_t \left( \Tr(\mathbf{F}_t^{\top} \hat{\mathbf{L}}_t \mathbf{F}_t) + \alpha \| \mathbf{F}_t - \mathbf{X}_t \mathbf{W}_t \|_F^2 \right) \\
& + \sum_{t=1}^{m} \langle \mathbf{Y}_t, \mathbf{W}_t - \mathbf{Z}_t \rangle  + \sum_{t=1}^{m} \frac{\rho}{2}\| \mathbf{W}_t - \mathbf{Z}_t \|_F^2 + \beta \| \bm{\mathcal{Z}} \|_{S_p}^p,\\
\end{aligned}
\end{equation}
where $\langle \cdot, \cdot \rangle$ denotes the inner product of Frobenius and $\rho > 0$ is the penalty parameter. The ADMM algorithm then proceeds by iteratively minimizing $L_{\rho}$ with respect to the primary variables $\{\mathbf{W}_t\}$ and $\bm{\mathcal{Z}}$, and then updating the dual variable tensor $\bm{\mathcal{Y}}$ by updating each slice $\mathbf{Y}_t$. \textit{(Note: The nonconvex orthogonality constraint $\mathbf{F}_t^{\top}\mathbf{F}_t=\mathbf{I}$ is handled directly within the subproblem for $\mathbf{F}_t$ and is thus not explicitly included in the Lagrangian.)}
\begin{algorithm}[t]
    \caption{Federated Multi-Task Clustering Framework}
    \label{alg:proposed_method_concise}
    \begin{algorithmic}[1]
        \STATE \textbf{Input:} Local datasets $\{\mathbf{X}_t\}_{t=1}^m$; number of clusters $k$; hyperparameters $\alpha, \beta, \rho$.
        \STATE \textbf{Output:} Converged local models $\{\mathbf{F}_t^*, \mathbf{W}_t^*\}$.
        \STATE \textbf{Initialization:}
        \STATE Server initializes global tensors $\bm{\mathcal{Z}}^0, \bm{\mathcal{Y}}^0$ to zero and weights $\lambda_t^0 \leftarrow 1/m$.
        \FOR{$t = 1$ to $m$ \textbf{in parallel}}
            \STATE Client $t$ computes its graph Laplacian $\hat{\mathbf{L}}_t$ and initializes local models $\mathbf{F}_t^0, \mathbf{W}_t^0$.
        \ENDFOR
        \WHILE{not converged}
            \STATE Server sends $\{\mathbf{Z}_t^k, \mathbf{Y}_t^k, \lambda_t^k\}$ to each client $t$.
            \FOR{$t = 1$ to $m$ \textbf{in parallel}}
                \STATE Update local projection matrix $\mathbf{W}_t^{k+1}$ per Eq. (12).
                \STATE Update local spectral embedding $\mathbf{F}_t^{k+1}$ per Eqs. (14)-(16).
                \STATE Client $t$ sends its updated $\mathbf{W}_t^{k+1}$ back to the server.
            \ENDFOR
            \STATE Server gathers $\{\mathbf{W}_t^{k+1}\}$ from all clients.
            \STATE Update the global consensus tensor $\bm{\mathcal{Z}}^{k+1}$ via TSVT per Eq. (18).
            \STATE Update the dual variables $\bm{\mathcal{Y}}^{k+1}$ per Eq. (19).
        \ENDWHILE
        \STATE \textbf{return} Final models $\{\mathbf{F}_t, \mathbf{W}_t\}$.
    \end{algorithmic}
\end{algorithm}
\subsection{Iterative Primal-Dual Update}
At each iteration $k+1$, the algorithm proceeds as follows.
\subsubsection{Primal Minimization Stage}
The primary variables, namely $\{\mathbf{F}_t, \mathbf{W}_t\}$ and the global tensor $\bm{\mathcal{Z}}$, are updated by minimizing the Augmented Lagrangian. We decompose this minimization into parallelizable updates on the clients and the server.

\textbf{Client-Side Updates:}
Each client $t$ is responsible for updating its local variables $\mathbf{W}_t$ and $\mathbf{F}_t$ through alternating minimization.

\textbf{a. Update of $\mathbf{W}_t$}: Fixing other variables, the subproblem for $\mathbf{W}_t$ is:
    \begin{equation}
    \begin{aligned} 
    \min_{\mathbf{W}_t} \Big\{ & \omega_t \alpha \| \mathbf{F}_t^k - \mathbf{X}_t \mathbf{W}_t \|_F^2 + \Tr((\mathbf{Y}_t^k)^{T}\mathbf{W}_t) \\
                              & + \frac{\rho}{2}\| \mathbf{W}_t - \mathbf{Z}_t^k \|_F^2 \Big\}.
    \end{aligned}
    \end{equation}
    As an unconstrained convex problem, we find the minimum by setting the partial derivative with respect to $\mathbf{W}_t$ to zero:
    \begin{equation}
        \begin{aligned}
        \frac{\partial L_\rho}{\partial \mathbf{W}_t} = &  \omega_t \alpha \cdot 2\mathbf{X}_t^{\top}(\mathbf{X}_t\mathbf{W}_t - \mathbf{F}_t^k) \\
        & + \mathbf{Y}_t^k + \rho(\mathbf{W}_t - \mathbf{Z}_t^k) = \mathbf{0},
        \end{aligned}
    \end{equation}
    Rearranging the terms to solve for $\mathbf{W}_t$ yields the following closed-form solution:
    \begin{equation}
        (2\omega_t\alpha \mathbf{X}_t^{\top} \mathbf{X}_t + \rho \mathbf{I}) \mathbf{W}_t = 2\omega_t\alpha \mathbf{X}_t^{\top} \mathbf{F}_t^k + \rho \mathbf{Z}_t^k - \mathbf{Y}_t^k,
    \end{equation}
\begin{equation}
\begin{aligned}
    \mathbf{W}_t^{k+1} ={} & (2\omega_t\alpha \mathbf{X}_t^{\top} \mathbf{X}_t + \rho \mathbf{I})^{-1} \\
    & \times (2\omega_t\alpha \mathbf{X}_t^{\top} \mathbf{F}_t^{k} + \rho \mathbf{Z}_t^k - \mathbf{Y}_t^k).
\end{aligned}
\end{equation}
\textbf{b. Update of $\mathbf{F}_t$}: With other variables fixed, the subproblem for $\mathbf{F}_t$ is:
    \begin{equation}
    \begin{gathered}
        \min_{\mathbf{F}_t} \quad \omega_t \left( \operatorname{Tr}(\mathbf{F}_t^{\top} \hat{\mathbf{L}}_t \mathbf{F}_t) + \alpha \| \mathbf{F}_t - \mathbf{X}_t \mathbf{W}_t^{k+1} \|_F^2 \right) \\
        \text{s.t.} \quad \mathbf{F}_t^{\top} \mathbf{F}_t = \mathbf{I}.
    \end{gathered}
    \end{equation}
    Leaving $\mathbf{A}_t = \mathbf{X}_t \mathbf{W}_t^{k+1}$, the objective for $\mathbf{F}_t$ becomes minimizing $f(\mathbf{F}_t) = \Tr(\mathbf{F}_t^{\top}(\hat{\mathbf{L}}_t + \alpha\mathbf{I})\mathbf{F}_t) - 2\alpha\Tr(\mathbf{F}_t^{\top} \mathbf{A}_t)$. This is an optimization problem on the Stiefel manifold and lacks a closed-form solution. We solve it by employing a projected gradient descent approach, which involves iteratively taking a step in the negative gradient direction followed by a projection back onto the manifold to maintain orthogonality. The Euclidean gradient of $f(\mathbf{F}_t)$ is:
    \begin{equation}
        \nabla_{\mathbf{F}_t} f = 2(\hat{\mathbf{L}}_t + \alpha\mathbf{I})\mathbf{F}_t - 2\alpha\mathbf{A}_t,
    \end{equation}
    Starting with $\mathbf{F}_t^{(0)} = \mathbf{F}_t^k$, we perform a few inner iterations. In each inner iteration $j$, we first take a gradient descent step with a step size $\eta$:
    \begin{equation}
        \mathbf{F}_{\text{cand}}^{(j+1)} = \mathbf{F}_t^{(j)} - \eta \left( (\hat{\mathbf{L}}_t + \alpha\mathbf{I})\mathbf{F}_t^{(j)} - \alpha\mathbf{A}_t \right).
    \end{equation}
    The resulting matrix $\mathbf{F}_{\text{cand}}^{(j+1)}$ is no longer guaranteed to be orthogonal. We enforce the constraint by projecting it back to the Stiefel manifold. This projection is achieved by finding the closest orthogonal matrix via Singular Value Decomposition (SVD). If the SVD of $\mathbf{F}_{\text{cand}}^{(j+1)}$ is $\mathbf{U \Sigma V}^{\top}$, the corrected orthogonal matrix is:
    \begin{equation}
        \mathbf{F}_t^{(j+1)} = \mathbf{U V}^{\top}.
    \end{equation}
    After a small number of such inner iterations, the final matrix is used as an update for the main ADMM loop, setting $\mathbf{F}_t^{k+1} = \mathbf{F}_t^{(j+1)}$.
\textbf{Server-Side Update.} The server, in turn, handles the global update for the consensus variable $\bm{\mathcal{Z}}$:
\begin{equation}
\begin{aligned}
\bm{\mathcal{Z}}^{k+1} = \operatorname*{arg\,min}_{\bm{\mathcal{Z}}} \bigg\{ \beta \| \bm{\mathcal{Z}} \|_{S_p}^p  & + \sum_{t=1}^{m} \Big( \operatorname{Tr}\left((\mathbf{Y}_t^k)^{\top}(-\mathbf{Z}_t)\right) \\
& + \frac{\rho}{2}\| \mathbf{W}_t^{k+1} - \mathbf{Z}_t \|_F^2 \Big) \bigg\}.
\end{aligned}
\end{equation}
By completing the square on the terms that involve $\bm{\mathcal{Z}}$, the problem can be simplified as follows:
\begin{equation}
\begin{aligned}
    \bm{\mathcal{Z}}^{k+1} = \argmin_{\bm{\mathcal{Z}}} \Big\{ & \beta \| \bm{\mathcal{Z}} \|_{S_p}^p + \\
    & \frac{\rho}{2}\| \bm{\mathcal{Z}} - (\bm{\mathcal{W}}^{k+1} + \frac{1}{\rho}\bm{\mathcal{Y}}^k) \|_F^2 \Big\}.
\end{aligned}
\end{equation}
This problem has a closed-form solution given by the Tensor Singular Value Thresholding (TSVT) \cite{lu2019tensor} algorithm.
\subsubsection{Dual Ascent Stage}
Finally, the dual variables $\bm{\mathcal{Y}}$ are updated through a gradient ascent step. This step leverages the primal residual $(\mathbf{W}_t^{k+1} - \mathbf{Z}_t^{k+1})$ to drive the local and global variables towards consensus.
\begin{equation}
    \mathbf{Y}_t^{k+1} = \mathbf{Y}_t^k + \rho (\mathbf{W}_t^{k+1} - \mathbf{Z}_t^{k+1}),
\end{equation}
where $\rho$ serves as the step size for the dual ascent. The dual variable tensor $\bm{\mathcal{Y}}$ thus accumulates the degree of constraint violation, providing a corrective signal in subsequent iterations that drives the primal variables towards consensus and, ultimately, to the optimal solution.

\subsection{Privacy and Security Considerations}
A core design principle of our FMTC framework is the preservation of data privacy at a fundamental level by avoiding the transmission of raw local data $X_t$. In our protocol, the only information transmitted from any client $C_t$ to the server is its updated local model parameter, the projection matrix $W_{t}^{k+1}$.

Sharing $W_{t}$ instead of $X_t$ provides a significant privacy enhancement over centralized methods, as it only reveals abstract patterns learned from the data. However, we acknowledge that this approach does not offer formal privacy guarantees against sophisticated model inversion attacks that aim to reconstruct data from shared model parameters. Our framework should be viewed as providing a baseline level of privacy that is common in many federated learning systems. For applications requiring rigorous privacy guarantees, FMTC could be integrated with advanced privacy-enhancing technologies such as differential privacy or secure aggregation in future work.

\section{Experimental Results}
\label{sec:experiments}

\begin{table*}[t]
\centering
\caption{Clustering performance comparison on various datasets. All metrics are in percent (\%). The best and second-best results across all methods are in \best{red bold} and \second{blue bold}, respectively.}
\label{tab:results_full_combined}

\label{tab:results_part1_combined}
\resizebox{\textwidth}{!}{%
\begin{tabular}{l|ccc|ccc||ccc|ccc|ccc}
\toprule
& \multicolumn{3}{c|}{\textbf{stSC\cite{ng2001spectral}}} & \multicolumn{3}{c||}{\textbf{uSC\cite{ng2001spectral}}} & \multicolumn{3}{c|}{\textbf{MBC\cite{zhang2011multitask}}} & \multicolumn{3}{c|}{\textbf{SMBC\cite{zhang2015smart}}} & \multicolumn{3}{c}{\textbf{SMKC\cite{zhang2015smart}}} \\
\textbf{Dataset} & \textbf{ACC} & \textbf{NMI} & \textbf{RI} & \textbf{ACC} & \textbf{NMI} & \textbf{RI} & \textbf{ACC} & \textbf{NMI} & \textbf{RI} & \textbf{ACC} & \textbf{NMI} & \textbf{RI} & \textbf{ACC} & \textbf{NMI} & \textbf{RI} \\
\midrule
WebKB4   & \cellcolor{lightflesh}63.37 & \cellcolor{lightflesh}21.44 & \cellcolor{lightflesh}62.45 & \cellcolor{lightflesh}63.76 & \cellcolor{lightflesh}20.75 & \cellcolor{lightflesh}62.56 & \cellcolor{lightblue}62.22 & \cellcolor{lightblue}18.86 & \cellcolor{lightblue}60.19 & \cellcolor{lightblue}67.05 & \cellcolor{lightblue}28.87 & \cellcolor{lightblue}66.40 & \cellcolor{lightblue}60.09 & \cellcolor{lightblue}12.06 & \cellcolor{lightblue}59.98 \\
20News   & \cellcolor{lightflesh}56.18 & \cellcolor{lightflesh}24.14 & \cellcolor{lightflesh}64.41 & \cellcolor{lightflesh}39.15 & \cellcolor{lightflesh}23.71 & \cellcolor{lightflesh}59.76 & \cellcolor{lightblue}44.75 & \cellcolor{lightblue}13.27 & \cellcolor{lightblue}47.43 & \cellcolor{lightblue}51.00 & \cellcolor{lightblue}22.96 & \cellcolor{lightblue}56.56 & \cellcolor{lightblue}36.09 & \cellcolor{lightblue}21.51 & \cellcolor{lightblue}59.50 \\
Reuters  & \cellcolor{lightflesh}\second{85.17} & \cellcolor{lightflesh}66.54 & \cellcolor{lightflesh}81.87 & \cellcolor{lightflesh}79.36 & \cellcolor{lightflesh}57.85 & \cellcolor{lightflesh}78.73 & \cellcolor{lightblue}75.71 & \cellcolor{lightblue}57.79 & \cellcolor{lightblue}75.71 & \cellcolor{lightblue}85.03 & \cellcolor{lightblue}\second{68.24} & \cellcolor{lightblue}\second{82.40} & \cellcolor{lightblue}69.93 & \cellcolor{lightblue}37.68 & \cellcolor{lightblue}69.12 \\
Keck     & \cellcolor{lightflesh}50.04 & \cellcolor{lightflesh}\second{48.80} & \cellcolor{lightflesh}83.25 & \cellcolor{lightflesh}47.32 & \cellcolor{lightflesh}44.85 & \cellcolor{lightflesh}82.60 & \cellcolor{lightblue}35.43 & \cellcolor{lightblue}25.98 & \cellcolor{lightblue}83.95 & \cellcolor{lightblue}37.95 & \cellcolor{lightblue}28.63 & \cellcolor{lightblue}84.24 & \cellcolor{lightblue}41.84 & \cellcolor{lightblue}31.08 & \cellcolor{lightblue}\second{84.57} \\
core50   & \cellcolor{lightflesh}39.63 & \cellcolor{lightflesh}5.43  & \cellcolor{lightflesh}\second{69.89} & \cellcolor{lightflesh}30.40 & \cellcolor{lightflesh}0.92  & \cellcolor{lightflesh}68.90 & \cellcolor{lightblue}40.27 & \cellcolor{lightblue}9.07  & \cellcolor{lightblue}63.63 & \cellcolor{lightblue}45.97 & \cellcolor{lightblue}15.87 & \cellcolor{lightblue}67.15 & \cellcolor{lightblue}27.28 & \cellcolor{lightblue}0.20  & \cellcolor{lightblue}62.55 \\
BBC News & \cellcolor{lightflesh}54.78 & \cellcolor{lightflesh}28.32 & \cellcolor{lightflesh}74.17 & \cellcolor{lightflesh}55.02 & \cellcolor{lightflesh}34.49 & \cellcolor{lightflesh}75.10 & \cellcolor{lightblue}68.17 & \cellcolor{lightblue}58.58 & \cellcolor{lightblue}82.29 & \cellcolor{lightblue}70.35 & \cellcolor{lightblue}56.48 & \cellcolor{lightblue}82.99 & \cellcolor{lightblue}74.22 & \cellcolor{lightblue}60.09 & \cellcolor{lightblue}84.87 \\
YALE     & \cellcolor{lightflesh}24.12 & \cellcolor{lightflesh}24.37 & \cellcolor{lightflesh}22.62 & \cellcolor{lightflesh}27.64 & \cellcolor{lightflesh}20.06 & \cellcolor{lightflesh}30.92 & \cellcolor{lightblue}34.48 & \cellcolor{lightblue}30.64 & \cellcolor{lightblue}35.16 & \cellcolor{lightblue}36.20 & \cellcolor{lightblue}39.94 & \cellcolor{lightblue}38.68 & \cellcolor{lightblue}45.33 & \cellcolor{lightblue}\second{41.15} & \cellcolor{lightblue}44.82 \\
\bottomrule
\end{tabular}%
}

\vspace{1.5em} 

\label{tab:results_part2_combined}
\resizebox{\textwidth}{!}{%
\begin{tabular}{l|ccc||ccc|ccc|ccc|ccc}
\toprule
& \multicolumn{3}{c||}{\textbf{MTSC\cite{yang2014multitask}}} & \multicolumn{3}{c|}{\textbf{SFOMVC\cite{feng2025scalable}}} & \multicolumn{3}{c|}{\textbf{Fed-kmeans\cite{garst2024federated}}} & \multicolumn{3}{c|}{\textbf{FedSpectral+\cite{thakkar2023fedspectral+}}} & \multicolumn{3}{c}{\textbf{Ours}} \\
\textbf{Dataset} & \textbf{ACC} & \textbf{NMI} & \textbf{RI} & \textbf{ACC} & \textbf{NMI} & \textbf{RI} & \textbf{ACC} & \textbf{NMI} & \textbf{RI} & \textbf{ACC} & \textbf{NMI} & \textbf{RI} & \textbf{ACC} & \textbf{NMI} & \textbf{RI} \\
\midrule
WebKB4   & \cellcolor{lightblue}\second{76.86} & \cellcolor{lightblue}\second{45.72} & \cellcolor{lightblue}\second{67.42} & \cellcolor{lightpink}73.65 & \cellcolor{lightpink}32.42 & \cellcolor{lightpink}57.62 & \cellcolor{lightpink}54.54 & \cellcolor{lightpink}15.91 & \cellcolor{lightpink}38.95 & \cellcolor{lightpink}50.86 & \cellcolor{lightpink}14.90 & \cellcolor{lightpink}61.12 & \cellcolor{lightpink}\best{78.11} & \cellcolor{lightpink}\best{46.04} & \cellcolor{lightpink}\best{74.28} \\
20News   & \cellcolor{lightblue}\best{72.32} & \cellcolor{lightblue}\second{65.90} & \cellcolor{lightblue}\second{80.73} & \cellcolor{lightpink}70.15 & \cellcolor{lightpink}62.55 & \cellcolor{lightpink}78.90 & \cellcolor{lightpink}30.35 & \cellcolor{lightpink}21.58 & \cellcolor{lightpink}29.47 & \cellcolor{lightpink}27.10 & \cellcolor{lightpink}25.27 & \cellcolor{lightpink}57.17 & \cellcolor{lightpink}\second{71.30} & \cellcolor{lightpink}\best{67.83} & \cellcolor{lightpink}\best{81.87} \\
Reuters  & \cellcolor{lightblue}77.89 & \cellcolor{lightblue}56.09 & \cellcolor{lightblue}73.00 & \cellcolor{lightpink}76.50 & \cellcolor{lightpink}54.20 & \cellcolor{lightpink}71.80 & \cellcolor{lightpink}69.37 & \cellcolor{lightpink}44.17 & \cellcolor{lightpink}66.98 & \cellcolor{lightpink}28.19 & \cellcolor{lightpink}15.14 & \cellcolor{lightpink}61.51 & \cellcolor{lightpink}\best{89.40} & \cellcolor{lightpink}\best{73.85} & \cellcolor{lightpink}\best{86.23} \\
Keck     & \cellcolor{lightblue}43.87 & \cellcolor{lightblue}38.85 & \cellcolor{lightblue}83.85 & \cellcolor{lightpink}48.22 & \cellcolor{lightpink}40.15 & \cellcolor{lightpink}82.50 & \cellcolor{lightpink}23.89 & \cellcolor{lightpink}23.27 & \cellcolor{lightpink}58.11 & \cellcolor{lightpink}\second{51.06} & \cellcolor{lightpink}42.91 & \cellcolor{lightpink}81.57 & \cellcolor{lightpink}\best{56.72} & \cellcolor{lightpink}\best{57.68} & \cellcolor{lightpink}\best{85.64} \\
core50   & \cellcolor{lightblue}\second{56.64} & \cellcolor{lightblue}20.38 & \cellcolor{lightblue}66.16 & \cellcolor{lightpink}54.30 & \cellcolor{lightpink}21.10 & \cellcolor{lightpink}65.50 & \cellcolor{lightpink}25.04 & \cellcolor{lightpink}0.45  & \cellcolor{lightpink}25.00 & \cellcolor{lightpink}47.32 & \cellcolor{lightpink}\second{22.87} & \cellcolor{lightpink}63.38 & \cellcolor{lightpink}\best{61.37} & \cellcolor{lightpink}\best{26.74} & \cellcolor{lightpink}\best{73.77} \\
BBC News & \cellcolor{lightblue}\second{76.67} & \cellcolor{lightblue}\second{63.67} & \cellcolor{lightblue}\second{84.90} & \cellcolor{lightpink}75.50 & \cellcolor{lightpink}62.90 & \cellcolor{lightpink}83.80 & \cellcolor{lightpink}22.96 & \cellcolor{lightpink}26.97 & \cellcolor{lightpink}24.36 & \cellcolor{lightpink}55.13 & \cellcolor{lightpink}22.22 & \cellcolor{lightpink}78.76 & \cellcolor{lightpink}\best{77.90} & \cellcolor{lightpink}\best{63.81} & \cellcolor{lightpink}\best{85.34} \\
YALE     & \cellcolor{lightblue}\second{54.10} & \cellcolor{lightblue}32.47 & \cellcolor{lightblue}\second{61.65} & \cellcolor{lightpink}52.80 & \cellcolor{lightpink}33.50 & \cellcolor{lightpink}60.10 & \cellcolor{lightpink}21.55 & \cellcolor{lightpink}18.21 & \cellcolor{lightpink}20.04 & \cellcolor{lightpink}49.81 & \cellcolor{lightpink}35.06 & \cellcolor{lightpink}55.19 & \cellcolor{lightpink}\best{60.35} & \cellcolor{lightpink}\best{48.12} & \cellcolor{lightpink}\best{67.28} \\
\bottomrule
\end{tabular}
}
\end{table*}

\subsection{Experimental Settings}

\paragraph{Datasets}
To ensure a comprehensive evaluation of the proposed FMTC framework, we conducted experiments on seven widely-used benchmark datasets: \textbf{WebKB4}, \textbf{20NewsGroups}, \textbf{Reuters}, \textbf{Keck}, \textbf{CORe50}\footnote{https://vlomonaco.github.io/core50/index.html}, \textbf{BBC News}\footnote{http://mlg.ucd.ie/datasets/bbc.html}, and \textbf{YALE}. These datasets span diverse modalities, ranging from high-dimensional sparse text documents to dense image vectors, allowing us to assess the method's versatility.

Crucially, to simulate a realistic cross-silo federated learning environment, we strictly follow a {Non-IID} data partitioning strategy. Instead of randomly shuffling data, we treat each natural data source as a distinct client (\emph{e.g.,} each university in WebKB4 or each news category in BBC acts as a client). This setting introduces significant statistical heterogeneity, as the data distribution varies across clients. 

\begin{figure}[!t]
    \centering
    \includegraphics[width=\columnwidth]{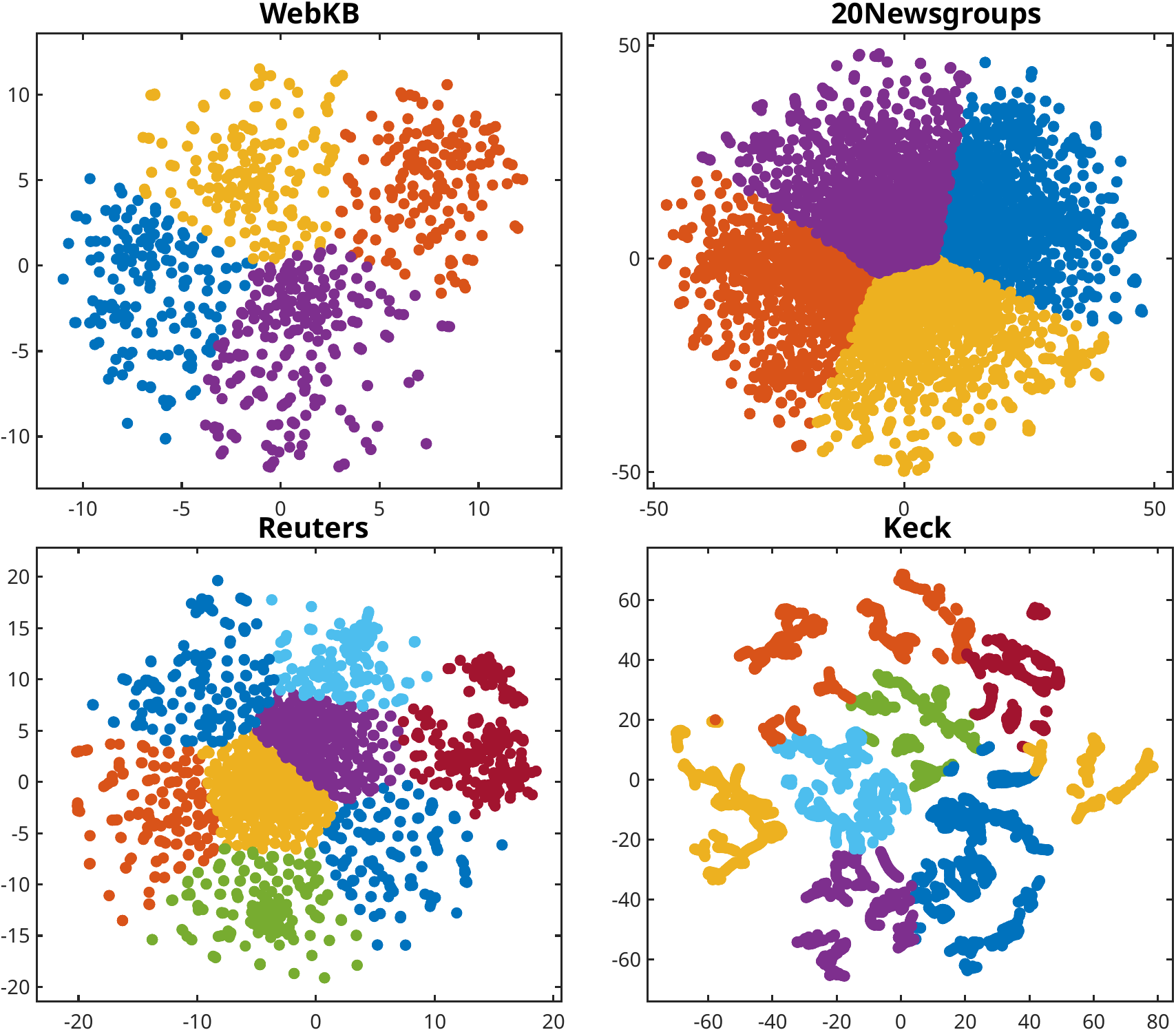} % 确保你的图片文件名为 web.pdf
    \caption{t-SNE visualization of the WebKB, 20Newsgroups, Reuters, and Keck datasets. Each color represents a distinct class. The plots highlight the significant heterogeneity and complex, non-linearly separable structures within and across tasks (clients), motivating the need for a personalized and collaborative clustering approach like FMTC.}
    \label{fig:tsne}
\end{figure}

As exemplified by the t-SNE visualizations in Fig.~\ref{fig:tsne} (WebKB, 20Newsgroups, Reuters, and Keck), the data points exhibit complex, non-linearly separable manifold structures. Moreover, the cluster boundaries differ visibly across tasks. This visualization underscores that a simple union of data or a single global model would fail to capture these fine-grained local structures, thereby motivating the need for our personalized clustering approach that can handle task-specific distributions while leveraging shared knowledge.

\paragraph{Compared Methods}

To rigorously evaluate the effectiveness of FMTC, we compare it against a comprehensive suite of baselines categorized into three groups. These methods are chosen to represent different levels of data access and collaboration strategies:

\begin{itemize}
    \item \textbf{Single-Task Baselines (stSC, uSC):} These methods represent the two extremes of data usage. \textbf{Spectral Clustering (stSC)} \cite{ng2001spectral} is trained independently on each client using only local data, serving as a baseline for the "isolated" scenario. Conversely, \textbf{Spectral Clustering-Union (uSC)} \cite{ng2001spectral} centrally pools all data into a single dataset, ignoring the distribution shift between clients and the multi-task structure.

    \item \textbf{Centralized Multi-Task Clustering Methods (MBC, SMBC, SMKC, MTSC):} These algorithms assume unrestricted access to raw data from all tasks and typically serve as the performance upper bound. We include:
    \textbf{Multi-task Bregman Clustering (MBC)} \cite{zhang2011multitask}, which minimizes an average Bregman divergence with task regularization;
    \textbf{Smart Multitask Bregman Clustering (SMBC)} \cite{zhang2015smart}, a transfer learning approach leveraging auxiliary data;
    \textbf{Smart Multi-task Kernel Clustering (SMKC)} \cite{zhang2015smart}, the kernelized variant of SMBC for non-linear structures;
    and \textbf{Multi-Task Spectral Clustering (MTSC)} \cite{yang2014multitask}, which explicitly captures task correlations via spectral analysis. Comparing against these methods validates whether FMTC can match centralized performance under privacy constraints.

    \item \textbf{Federated Clustering Methods (Fed-kmeans, FedSpectral+, SFOMVC, Ours):} These are the direct competitors in the privacy-preserving setting. \textbf{Federated K-Means (Fed-kmeans)} \cite{garst2024federated} aggregates local centroids; \textbf{Federated Spectral Clustering (FedSpectral+)} \cite{thakkar2023fedspectral+} aggregates Laplacian eigenvectors; and \textbf{Scalable Federated Orthogonal Multi-View Clustering (SFOMVC)} \cite{feng2025scalable} handles multi-view data in federated settings. However, these methods typically employ a ``one-model-fits-all'' strategy or simple parameter averaging, often failing to account for the personalized manifold structures of heterogeneous clients. Our proposed \textbf{FMTC} addresses these limitations by learning personalized models while explicitly capturing inter-client correlations through tensor-based regularization.
\end{itemize}

\paragraph{Evaluation Metrics.}
By following standard protocols in spectral clustering, we perform k-means on the normalized rows of the learned embedding matrices $\{\mathbf{F}_t\}$ to obtain discrete cluster assignments. To mitigate the sensitivity of k-means to initialization, we repeat the clustering process 10 times with different random seeds and report the result with the lowest minimal sum of squared errors. The clustering quality is quantitatively evaluated with three standard metrics: Clustering Accuracy (ACC), Normalized Mutual Information (NMI), and Rand Index (RI)\cite{manning2008introduction}. ACC is computed using the Hungarian algorithm to find the optimal one-to-one mapping between cluster assignments and ground-truth labels. Higher values of all three metrics indicate better clustering performance.

\begin{figure*}[htbp] 
    \centering
    \includegraphics[width=\textwidth]{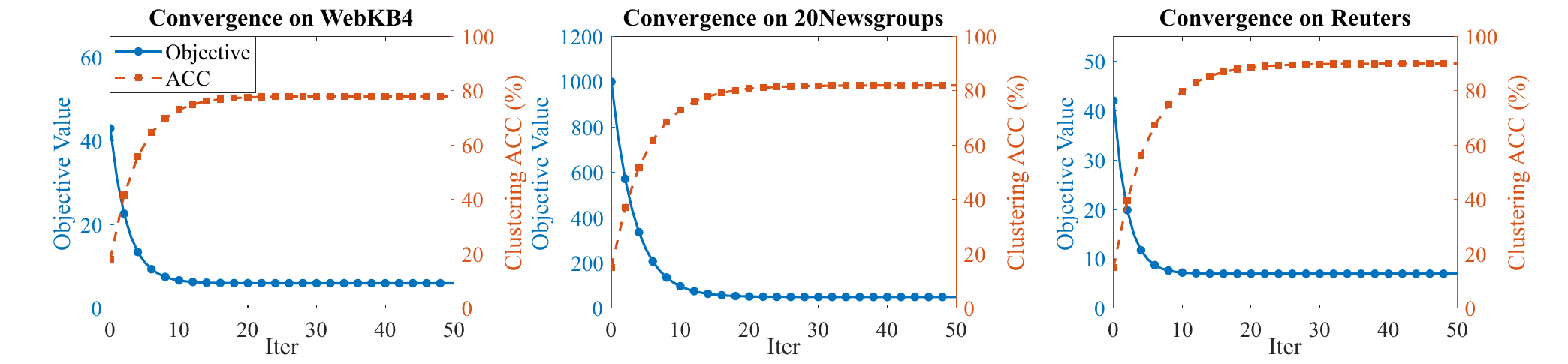} 
    \caption{Convergence analysis of our proposed FMTC framework on three datasets. For each dataset, we plot the overall objective function value and the average clustering ACC (\%) versus the number of iterations.}
    \label{fig:convergence}
\end{figure*}

\begin{figure}[t]
    \centering
    \includegraphics[width=\columnwidth]{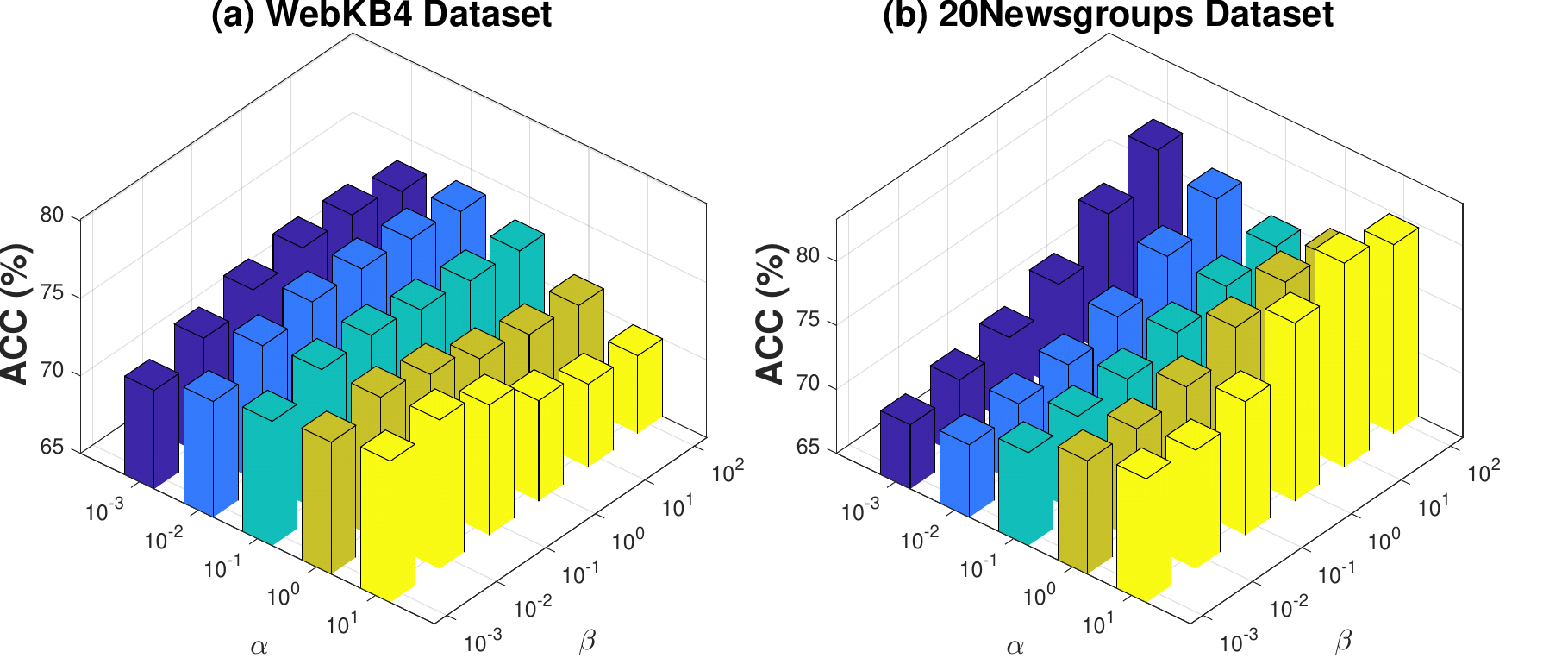}
    \caption{Parameter sensitivity analysis of FMTC on the WebKB4 dataset. The figure shows the ACC (\%) as a function of the collaboration weight $\beta$ and the local fidelity weight $\alpha$. Each row, corresponding to a specific $\alpha$, is rendered in a unique color.}
    \label{fig:sensitivity_bar3}
\end{figure}

\subsection{Experiment Results}

The quantitative clustering results on all seven datasets are summarized in Table~\ref{tab:results_full_combined}. From the reported results, we have the following observations:

\begin{itemize}
    \item \textbf{Compared with single-task spectral clustering methods}, our proposed FMTC can obtain more than 14.35, 20.74, 7.13 and 14.12 percent improvements on WebKB4, CORe50, Reuters and 20NewsGroups datasets in terms of Avg.ACC, respectively. It is because FMTC could utilize the relationships among multiple clustering tasks, whereas the single spectral clustering methods (i.e., stSC and uSC) implement each task independently or pool all data together while ignoring the multi-task structure. More specifically, on the highly heterogeneous 20NewsGroups dataset, FMTC achieves 71.30\% in ACC, significantly outperforming stSC (56.18\%) and uSC (39.15\%). This demonstrates the effectiveness of capturing inter-client correlations in federated settings.
    
    \item \textbf{Compared with centralized multi-task clustering methods}, we could see that FMTC outperforms these competing methods in most cases despite not having direct access to raw data from all clients. This observation verifies the effectiveness of learning the latent cluster centers between each pair of spectral tasks and preserving the common embedded features across multiple clustering tasks. Notably, FMTC consistently outperforms MTSC, a centralized method with access to raw data. On the WebKB4 dataset, FMTC improves NMI by approximately 0.32\% over MTSC. This result empirically validates our hypothesis that the tensor low-rank regularization (Schatten $p$-norm) is more effective than traditional matrix-based regularization used in MTSC. Matrix-based methods can only capture pairwise task correlations by flattening the model structure, whereas FMTC explicitly captures the high-order, latent semantic relationships among tasks by organizing client models into a third-order tensor.
    
    \item \textbf{Compared with federated clustering methods}, FMTC significantly outperforms Fed-kmeans and FedSpectral+ across all datasets, particularly on highly heterogeneous data. For instance, on 20NewsGroups, Fed-kmeans and FedSpectral+ achieve only 30.35\% and 27.10\% accuracy, respectively, whereas FMTC achieves a substantial improvement to 71.30\%. This drastic performance gap highlights a fundamental limitation of current federated clustering approaches: they typically rely on simple parameter averaging or centroid aggregation. In Non-IID scenarios where the underlying data distributions diverge significantly across clients, such averaging operations wash out the distinct manifold structures learned by local clients, leading to a collapsed global model that fits no one. In sharp contrast, FMTC maintains personalized projection matrices $\{\mathbf{W}_t\}$ for each client, preserving their unique local structures, while using the global tensor $\bm{\mathcal{Z}}$ only as a regularizer to guide the learning without enforcing strict homogeneity.

\end{itemize}

\subsection{Convergence Analysis}
We provide a rigorous theoretical proof of convergence for our algorithm in the appendix, which guarantees that the iterates converge to a KKT stationary point by leveraging established results for non-convex ADMM\cite{wang2019global}. To complement this theoretical analysis and validate the practical effectiveness and stability of our proposed ADMM-based optimization algorithm, we visualize its convergence behavior on three representative datasets, WebKB4, 20NewsGroups and Reuters. Figure~\ref{fig:convergence} plots the value of the overall objective function and the average clustering ACC against the number of iterations. As can be observed, the proposed FMTC framework exhibits excellent convergence properties across all tasks. In each case, the objective function value decreases monotonically and stationarity, stabilizing after approximately 15-20 iterations. Concurrently, the clustering ACC shows a steep increase in the initial iterations and then converges smoothly to a high-performance plateau. This fast and stable convergence demonstrates that our algorithm can efficiently find a high-quality solution in a limited number of communication rounds, making it practical for real-world federated learning scenarios.

\begin{table*}[t]
\centering
\caption{Comparison of In-Sample (IS) and Out-of-Sample (OOS) clustering performance (\%). For each method, OOS results are obtained on a held-out 20\% test set. The gap between IS and OOS performance indicates the generalization ability. The best and second-best results across all methods are in \best{red bold} and \second{blue bold}, respectively.}
\label{tab:oos_results_final}
\resizebox{\textwidth}{!}{%
\begin{tabular}{l|cc|cc||cc|cc|cc|cc||cc|cc|cc}

\toprule
& \multicolumn{2}{c|}{\textbf{stSC\cite{ng2001spectral}}} & \multicolumn{2}{c||}{\textbf{uSC\cite{ng2001spectral}}} & \multicolumn{2}{c|}{\textbf{MBC\cite{zhang2011multitask}}} & \multicolumn{2}{c|}{\textbf{SMBC\cite{zhang2015smart}}} & \multicolumn{2}{c|}{\textbf{SMKC\cite{zhang2015smart}}} & \multicolumn{2}{c||}{\textbf{MTSC\cite{yang2014multitask}}} & \multicolumn{2}{c|}{\textbf{Fed-kmeans\cite{garst2024federated}}} & \multicolumn{2}{c|}{\textbf{FedSpectral+\cite{thakkar2023fedspectral+}}} & \multicolumn{2}{c}{\textbf{Ours}} \\
\cmidrule(lr){2-3} \cmidrule(lr){4-5} \cmidrule(lr){6-7} \cmidrule(lr){8-9} \cmidrule(lr){10-11} \cmidrule(lr){12-13} \cmidrule(lr){14-15} \cmidrule(lr){16-17} \cmidrule(lr){18-19}
\textbf{Dataset} & \textbf{IS} & \textbf{OOS} & \textbf{IS} & \textbf{OOS} & \textbf{IS} & \textbf{OOS} & \textbf{IS} & \textbf{OOS} & \textbf{IS} & \textbf{OOS} & \textbf{IS} & \textbf{OOS} & \textbf{IS} & \textbf{OOS} & \textbf{IS} & \textbf{OOS} & \textbf{IS} & \textbf{OOS} \\
\midrule
\multicolumn{19}{c}{\textbf{Metric: ACC (\%)}} \\
\midrule
WebKB4 & \cellcolor{lightflesh}63.37 & \cellcolor{lightflesh}49.43 & \cellcolor{lightflesh}63.76 & \cellcolor{lightflesh}49.10 & \cellcolor{lightblue}62.22 & \cellcolor{lightblue}46.67 & \cellcolor{lightblue}67.05 & \cellcolor{lightblue}53.64 & \cellcolor{lightblue}60.09 & \cellcolor{lightblue}43.26 & \cellcolor{lightblue}\second{76.86} & \cellcolor{lightblue}\second{62.26} & \cellcolor{lightpink}54.54 & \cellcolor{lightpink}40.36 & \cellcolor{lightpink}50.86 & \cellcolor{lightpink}35.60 & \cellcolor{lightpink}\best{78.11} & \cellcolor{lightpink}\best{64.67} \\
20News & \cellcolor{lightflesh}56.18 & \cellcolor{lightflesh}49.44 & \cellcolor{lightflesh}39.15 & \cellcolor{lightflesh}33.28 & \cellcolor{lightblue}44.75 & \cellcolor{lightblue}38.48 & \cellcolor{lightblue}51.00 & \cellcolor{lightblue}46.92 & \cellcolor{lightblue}36.09 & \cellcolor{lightblue}30.32 & \cellcolor{lightblue}\best{72.32} & \cellcolor{lightblue}\second{68.70} & \cellcolor{lightpink}30.35 & \cellcolor{lightpink}24.89 & \cellcolor{lightpink}27.10 & \cellcolor{lightpink}21.68 & \cellcolor{lightpink}\second{71.30} & \cellcolor{lightpink}\best{79.16} \\
Reuters & \cellcolor{lightflesh}\second{85.17} & \cellcolor{lightflesh}78.36 & \cellcolor{lightflesh}79.36 & \cellcolor{lightflesh}72.22 & \cellcolor{lightblue}75.71 & \cellcolor{lightblue}68.14 & \cellcolor{lightblue}85.03 & \cellcolor{lightblue}\second{79.93} & \cellcolor{lightblue}69.93 & \cellcolor{lightblue}61.54 & \cellcolor{lightblue}77.89 & \cellcolor{lightblue}72.44 & \cellcolor{lightpink}69.37 & \cellcolor{lightpink}61.74 & \cellcolor{lightpink}28.19 & \cellcolor{lightpink}22.55 & \cellcolor{lightpink}\best{89.40} & \cellcolor{lightpink}\best{85.27} \\
Keck & \cellcolor{lightflesh}50.04 & \cellcolor{lightflesh}33.53 & \cellcolor{lightflesh}47.32 & \cellcolor{lightflesh}31.23 & \cellcolor{lightblue}35.43 & \cellcolor{lightblue}23.03 & \cellcolor{lightblue}37.95 & \cellcolor{lightblue}25.81 & \cellcolor{lightblue}41.84 & \cellcolor{lightblue}27.61 & \cellcolor{lightblue}43.87 & \cellcolor{lightblue}31.13 & \cellcolor{lightpink}23.89 & \cellcolor{lightpink}15.77 & \cellcolor{lightpink}\second{51.06} & \cellcolor{lightpink}\second{34.21} & \cellcolor{lightpink}\best{56.72} & \cellcolor{lightpink}\best{47.64} \\
core50 & \cellcolor{lightflesh}39.63 & \cellcolor{lightflesh}28.14 & \cellcolor{lightflesh}30.40 & \cellcolor{lightflesh}20.98 & \cellcolor{lightblue}40.27 & \cellcolor{lightblue}28.19 & \cellcolor{lightblue}45.97 & \cellcolor{lightblue}33.56 & \cellcolor{lightblue}27.28 & \cellcolor{lightblue}18.55 & \cellcolor{lightblue}\second{56.64} & \cellcolor{lightblue}\second{41.35} & \cellcolor{lightpink}25.04 & \cellcolor{lightpink}17.03 & \cellcolor{lightpink}47.32 & \cellcolor{lightpink}32.65 & \cellcolor{lightpink}\best{61.37} & \cellcolor{lightpink}\best{45.36} \\
BBC News & \cellcolor{lightflesh}54.78 & \cellcolor{lightflesh}48.21 & \cellcolor{lightflesh}55.02 & \cellcolor{lightflesh}47.87 & \cellcolor{lightblue}68.17 & \cellcolor{lightblue}59.31 & \cellcolor{lightblue}70.35 & \cellcolor{lightblue}62.61 & \cellcolor{lightblue}74.22 & \cellcolor{lightblue}65.31 & \cellcolor{lightblue}\second{76.67} & \cellcolor{lightblue}\second{69.77} & \cellcolor{lightpink}22.96 & \cellcolor{lightpink}18.83 & \cellcolor{lightpink}55.13 & \cellcolor{lightpink}46.31 & \cellcolor{lightpink}\best{77.90} & \cellcolor{lightpink}\best{74.40} \\
\midrule
\multicolumn{19}{c}{\textbf{Metric: NMI (\%)}} \\
\midrule
WebKB4 & \cellcolor{lightflesh}21.44 & \cellcolor{lightflesh}13.29 & \cellcolor{lightflesh}20.75 & \cellcolor{lightflesh}12.45 & \cellcolor{lightblue}18.86 & \cellcolor{lightblue}10.94 & \cellcolor{lightblue}28.87 & \cellcolor{lightblue}18.77 & \cellcolor{lightblue}12.06 & \cellcolor{lightblue}6.63 & \cellcolor{lightblue}\second{45.72} & \cellcolor{lightblue}\second{31.09} & \cellcolor{lightpink}15.91 & \cellcolor{lightpink}9.07 & \cellcolor{lightpink}14.90 & \cellcolor{lightpink}7.45 & \cellcolor{lightpink}\best{46.04} & \cellcolor{lightpink}\best{32.32} \\
20News & \cellcolor{lightflesh}24.14 & \cellcolor{lightflesh}19.31 & \cellcolor{lightflesh}23.71 & \cellcolor{lightflesh}18.49 & \cellcolor{lightblue}13.27 & \cellcolor{lightblue}9.95 & \cellcolor{lightblue}22.96 & \cellcolor{lightblue}18.83 & \cellcolor{lightblue}21.51 & \cellcolor{lightblue}16.35 & \cellcolor{lightblue}\second{65.90} & \cellcolor{lightblue}\second{56.67} & \cellcolor{lightpink}21.58 & \cellcolor{lightpink}16.18 & \cellcolor{lightpink}25.27 & \cellcolor{lightpink}18.20 & \cellcolor{lightpink}\best{67.83} & \cellcolor{lightpink}\best{60.19} \\
Reuters & \cellcolor{lightflesh}66.54 & \cellcolor{lightflesh}59.89 & \cellcolor{lightflesh}57.85 & \cellcolor{lightflesh}50.91 & \cellcolor{lightblue}57.79 & \cellcolor{lightblue}49.70 & \cellcolor{lightblue}\second{68.24} & \cellcolor{lightblue}\second{62.78} & \cellcolor{lightblue}37.68 & \cellcolor{lightblue}30.90 & \cellcolor{lightblue}56.09 & \cellcolor{lightblue}51.04 & \cellcolor{lightpink}44.17 & \cellcolor{lightpink}37.10 & \cellcolor{lightpink}15.14 & \cellcolor{lightpink}33.86 & \cellcolor{lightpink}\best{73.85} & \cellcolor{lightpink}\best{69.88} \\
Keck & \cellcolor{lightflesh}\second{48.80} & \cellcolor{lightflesh}\second{32.21} & \cellcolor{lightflesh}44.85 & \cellcolor{lightflesh}29.15 & \cellcolor{lightblue}25.98 & \cellcolor{lightblue}16.63 & \cellcolor{lightblue}28.63 & \cellcolor{lightblue}18.89 & \cellcolor{lightblue}31.08 & \cellcolor{lightblue}20.20 & \cellcolor{lightblue}38.85 & \cellcolor{lightblue}27.97 & \cellcolor{lightpink}23.27 & \cellcolor{lightpink}15.12 & \cellcolor{lightpink}42.91 & \cellcolor{lightpink}27.89 & \cellcolor{lightpink}\best{57.68} & \cellcolor{lightpink}\best{47.85} \\
core50 & \cellcolor{lightflesh}5.43 & \cellcolor{lightflesh}3.53 & \cellcolor{lightflesh}0.92 & \cellcolor{lightflesh}0.55 & \cellcolor{lightblue}9.07 & \cellcolor{lightblue}5.80 & \cellcolor{lightblue}15.87 & \cellcolor{lightblue}10.47 & \cellcolor{lightblue}0.20 & \cellcolor{lightblue}0.11 & \cellcolor{lightblue}20.38 & \cellcolor{lightblue}\second{14.06} & \cellcolor{lightpink}0.45 & \cellcolor{lightpink}0.28 & \cellcolor{lightpink}\second{22.87} & \cellcolor{lightpink}\best{15.32} & \cellcolor{lightpink}\best{26.74} & \cellcolor{lightpink}8.82 \\
BBC News & \cellcolor{lightflesh}28.32 & \cellcolor{lightflesh}24.07 & \cellcolor{lightflesh}34.49 & \cellcolor{lightflesh}28.97 & \cellcolor{lightblue}58.58 & \cellcolor{lightblue}51.55 & \cellcolor{lightblue}56.48 & \cellcolor{lightblue}50.27 & \cellcolor{lightblue}60.09 & \cellcolor{lightblue}53.48 & \cellcolor{lightblue}\second{63.67} & \cellcolor{lightblue}\best{58.58} & \cellcolor{lightpink}26.97 & \cellcolor{lightpink}22.12 & \cellcolor{lightpink}22.22 & \cellcolor{lightpink}18.22 & \cellcolor{lightpink}\best{63.81} & \cellcolor{lightpink}\second{58.21} \\
\midrule
\multicolumn{19}{c}{\textbf{Metric: RI (\%)}} \\
\midrule
WebKB4 & \cellcolor{lightflesh}62.45 & \cellcolor{lightflesh}47.46 & \cellcolor{lightflesh}62.56 & \cellcolor{lightflesh}46.92 & \cellcolor{lightblue}60.19 & \cellcolor{lightblue}43.34 & \cellcolor{lightblue}66.40 & \cellcolor{lightblue}52.46 & \cellcolor{lightblue}59.98 & \cellcolor{lightblue}41.99 & \cellcolor{lightblue}\second{67.42} & \cellcolor{lightblue}\second{53.94} & \cellcolor{lightpink}38.95 & \cellcolor{lightpink}25.32 & \cellcolor{lightpink}61.12 & \cellcolor{lightpink}45.23 & \cellcolor{lightpink}\best{74.28} & \cellcolor{lightpink}\best{60.99} \\
20News & \cellcolor{lightflesh}64.41 & \cellcolor{lightflesh}57.97 & \cellcolor{lightflesh}59.76 & \cellcolor{lightflesh}52.59 & \cellcolor{lightblue}47.43 & \cellcolor{lightblue}40.32 & \cellcolor{lightblue}56.56 & \cellcolor{lightblue}52.04 & \cellcolor{lightblue}59.50 & \cellcolor{lightblue}51.17 & \cellcolor{lightblue}\second{80.73} & \cellcolor{lightblue}\best{78.31} & \cellcolor{lightpink}29.47 & \cellcolor{lightpink}23.58 & \cellcolor{lightpink}57.17 & \cellcolor{lightpink}48.60 & \cellcolor{lightpink}\best{81.87} & \cellcolor{lightpink}\second{71.30} \\
Reuters & \cellcolor{lightflesh}81.87 & \cellcolor{lightflesh}75.32 & \cellcolor{lightflesh}78.73 & \cellcolor{lightflesh}71.64 & \cellcolor{lightblue}75.71 & \cellcolor{lightblue}67.38 & \cellcolor{lightblue}\second{82.40} & \cellcolor{lightblue}\second{77.46} & \cellcolor{lightblue}69.12 & \cellcolor{lightblue}60.14 & \cellcolor{lightblue}73.00 & \cellcolor{lightblue}67.89 & \cellcolor{lightpink}66.98 & \cellcolor{lightpink}58.94 & \cellcolor{lightpink}61.51 & \cellcolor{lightpink}52.28 & \cellcolor{lightpink}\best{86.23} & \cellcolor{lightpink}\best{82.35} \\
Keck & \cellcolor{lightflesh}83.25 & \cellcolor{lightflesh}71.59 & \cellcolor{lightflesh}82.60 & \cellcolor{lightflesh}70.21 & \cellcolor{lightblue}83.95 & \cellcolor{lightblue}72.20 & \cellcolor{lightblue}84.24 & \cellcolor{lightblue}73.29 & \cellcolor{lightblue}84.57 & \cellcolor{lightblue}73.58 & \cellcolor{lightblue}83.85 & \cellcolor{lightblue}\second{74.63} & \cellcolor{lightpink}58.11 & \cellcolor{lightpink}48.23 & \cellcolor{lightpink}81.57 & \cellcolor{lightpink}69.33 & \cellcolor{lightpink}\best{85.64} & \cellcolor{lightpink}\best{78.25} \\
core50 & \cellcolor{lightflesh}\second{69.89} & \cellcolor{lightflesh}\second{60.10} & \cellcolor{lightflesh}68.90 & \cellcolor{lightflesh}58.56 & \cellcolor{lightblue}63.63 & \cellcolor{lightblue}53.45 & \cellcolor{lightblue}67.15 & \cellcolor{lightblue}57.75 & \cellcolor{lightblue}62.55 & \cellcolor{lightblue}52.54 & \cellcolor{lightblue}66.16 & \cellcolor{lightblue}58.22 & \cellcolor{lightpink}25.00 & \cellcolor{lightpink}19.50 & \cellcolor{lightpink}63.38 & \cellcolor{lightpink}54.51 & \cellcolor{lightpink}\best{73.77} & \cellcolor{lightpink}\best{69.89} \\
BBC News & \cellcolor{lightflesh}74.17 & \cellcolor{lightflesh}68.22 & \cellcolor{lightflesh}75.10 & \cellcolor{lightflesh}68.34 & \cellcolor{lightblue}82.29 & \cellcolor{lightblue}75.71 & \cellcolor{lightblue}82.99 & \cellcolor{lightblue}77.18 & \cellcolor{lightblue}84.87 & \cellcolor{lightblue}78.93 & \cellcolor{lightblue}\second{84.90} & \cellcolor{lightblue}\best{79.81} & \cellcolor{lightpink}24.36 & \cellcolor{lightpink}20.22 & \cellcolor{lightpink}78.76 & \cellcolor{lightpink}71.26 & \cellcolor{lightpink}\best{85.34} & \cellcolor{lightpink}\second{79.54} \\
\bottomrule
\end{tabular}%
}
\end{table*}

\begin{figure*}[!t]
    \centering
    \includegraphics[width=\textwidth]{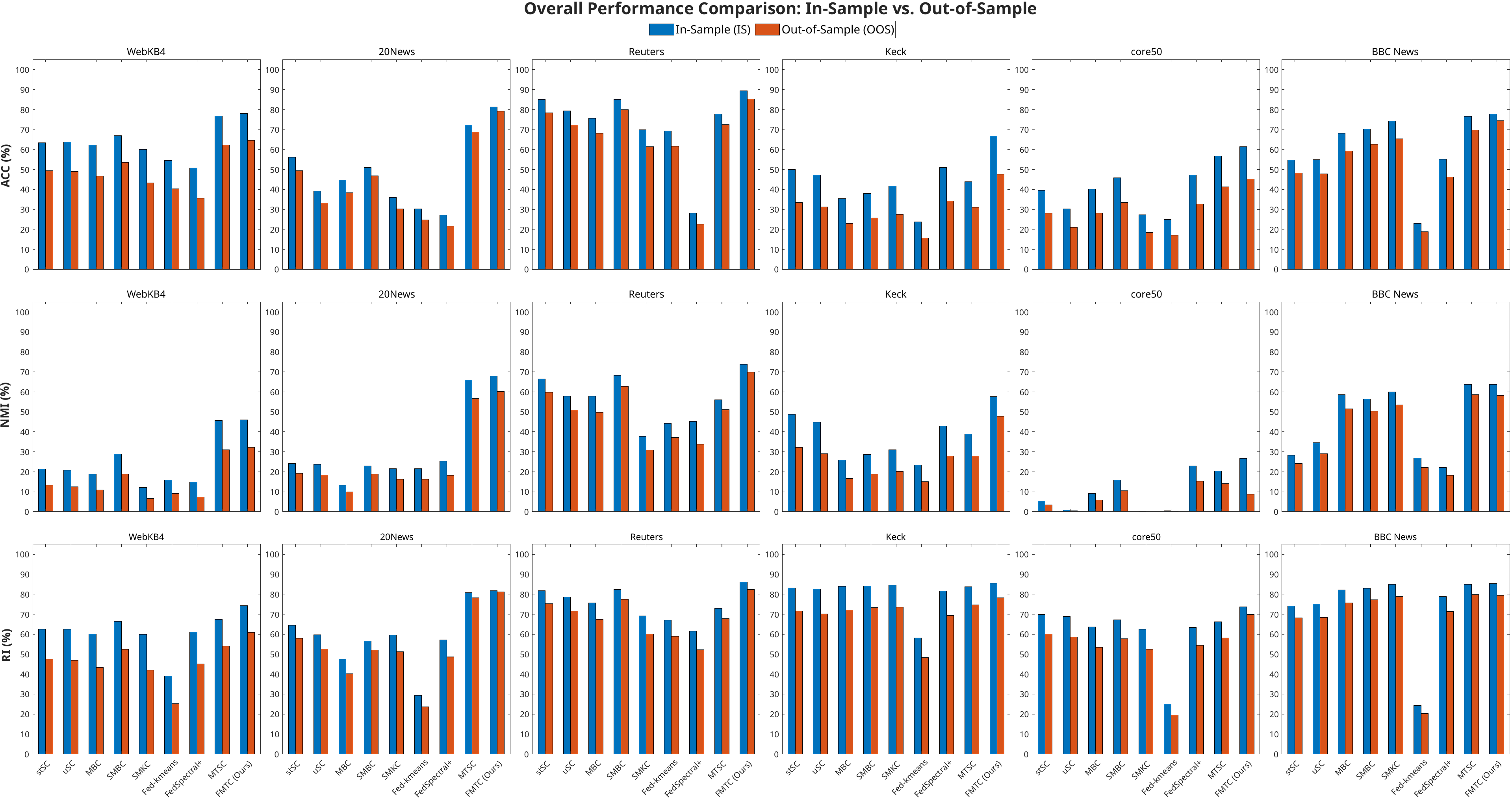}
    \caption{Comprehensive comparison of In-Sample (IS) and Out-of-Sample (OOS) performance across all six datasets and three evaluation metrics (ACC, NMI, RI). Each subplot displays the performance of all nine competing methods on a specific task. The blue bars represent the performance on the training data (IS), while the orange bars represent the performance on the held-out test data (OOS). The gap between the blue and orange bars visually represents the generalization gap for each method. It is consistently observed that our proposed FMTC method not only achieves the highest performance but also exhibits one of the smallest generalization gaps, highlighting its superior robustness and ability to generalize to unseen data.}
    \label{fig:is_vs_oos_comparison}
\end{figure*}

\subsection{Hyperparameter Sensitivity Analysis}

To investigate the impact of our model's key hyperparameters, we analyze the performance sensitivity with respect to the local fidelity weight $\alpha$ and the global collaboration weight $\beta$ on two representative datasets. The results are visualized in Fig.~\ref{fig:sensitivity_bar3}. The performance landscapes are shown to be complex and non-convex, with multiple ridges and local optima, reflecting the intricate interplay of the hyperparameters in real-world scenarios. For the WebKB4 dataset (Fig.~\ref{fig:sensitivity_bar3}a), a high-performance ridge is observed where the collaboration weight $\beta$ is moderate (around $10^{-1}$ to $10^{0}$). In contrast, the more complex 20Newsgroups dataset (Fig.~\ref{fig:sensitivity_bar3}b) exhibits a broader optimal region at a higher collaboration level ($\beta \approx 10^{1}$), suggesting a greater benefit from inter-task knowledge sharing. This demonstrates that while the optimal settings are dataset-dependent, the model is robust, maintaining strong performance across continuous regions rather than at sharp, isolated points. Importantly, for both datasets, performance consistently degrades as $\beta$ approaches zero, empirically validating the necessity of our server-side tensorial correlation module.

\subsection{Out-of-Sample Extension Analysis}

Existing methods that attempt to extend spectral clustering often resort to a disjoint "two-stage" approach: first generating pseudo-labels via clustering, and then training a separate classifier. As evidenced by the sharp performance drop of {FedSpectral+} on unseen data (\emph{e.g.}, a drop of over 20\% on Reuters), this decoupled strategy suffers from severe {error propagation}—noise in the initial clustering is memorized by the classifier, leading to poor generalization.
In contrast, FMTC demonstrates superior robustness with a minimal generalization gap. On the Reuters dataset, the OOS accuracy remains impressively high at {85.27\%}, significantly surpassing all baselines. This success is attributed to our novel {coupled optimization objective} (Eq. 4). By simultaneously minimizing the spectral cut and the projection error $\|\mathbf{F}_t - \mathbf{X}_t \mathbf{W}_t\|_F^2$, we create a closed-loop system where the mapping function $\mathbf{W}_t$ is not just a passive learner. Instead, it acts as a {structural regularizer}, forcing the spectral embeddings $\mathbf{F}_t$ to lie in a subspace that is predictable by linear features. This "predictability constraint" effectively smoothes the learned manifold, preventing the model from overfitting to the idiosyncrasies of specific training points and ensuring that the learned structure is valid for the underlying data distribution.
The ability to maintain high performance on OOS data confirms that our FMTC is ready for practical deployment. Once trained, the lightweight projection matrices $\{\mathbf{W}_t\}$ can be frozen and deployed on edge devices for real-time inference on streaming data, eliminating the need for frequent and costly model retraining.

%%%%%%%%%%%%%%%%%%%%%%%%%%%%%%%%%%%%%%%%%%%%%%%%%%%%%%%%%%%%%%%%%%%%%

\section{Conclusion}
\label{sec:conclusion}

In this work, we present a \underline{F}ederated \underline{M}ulti-\underline{T}ask \underline{C}lustering (\emph{i.e.,} {FMTC}) model, a novel framework that addresses the dual challenges of intra-client correlation and inter-client correlation. The FMTC framework introduces two key components: client-side personalized clustering module and server-side tensorial correlation module. The {client-side personalized clustering module} integrates spectral embedding learning with parametric mapping function optimization into a unified closed-loop system, eliminating pseudo-label dependency and enabling out-of-sample inference. The {server-side tensorial correlation module} organizes local client models into a third-order tensor with low-rank regularization, explicitly capturing high-order shared knowledge while preserving personalized local structures. These two modules are jointly optimized via an efficient ADMM-based algorithm that ensures privacy preservation. Extensive experiments across seven benchmark datasets demonstrate that FMTC significantly outperforms state-of-the-art federated clustering algorithms, particularly in scenarios requiring strong generalization to unseen data. For future work, addressing the rotational ambiguity of learned projection matrices could further enhance performance.
%%%%%%%%%%%%%%%%%%%%%%%%%%%%%%%%%%%%%%%%%%%%%%%%%%%%%%%%%

\bibliographystyle{IEEEtran}
\bibliography{REF}

\begin{IEEEbiography}[{\includegraphics[width=1in,height=1.25in,clip,keepaspectratio]{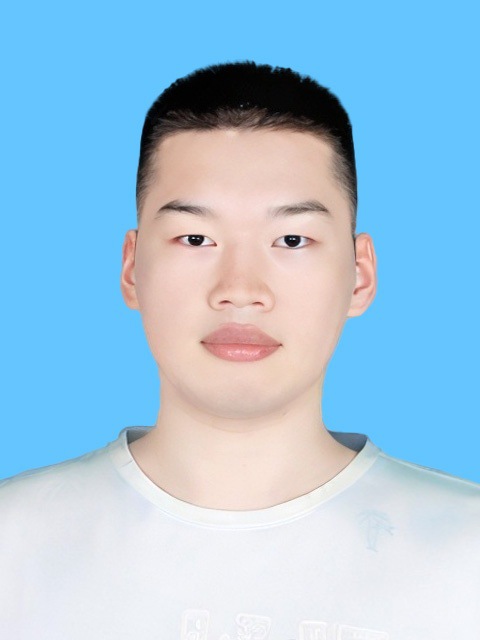}}]{Suyan Dai}
received his B.S. degree from the Shien-Ming Wu School of Intelligent Engineering, South China University of Technology, China, in 2024, where he is currently pursuing the M.S. degree with the School of Automation Science and Engineering. His research interests include robotics, reinforcement learning, and deep learning.
\end{IEEEbiography}

\begin{IEEEbiography}[{\includegraphics[width=1in,height=1.25in,clip,keepaspectratio]{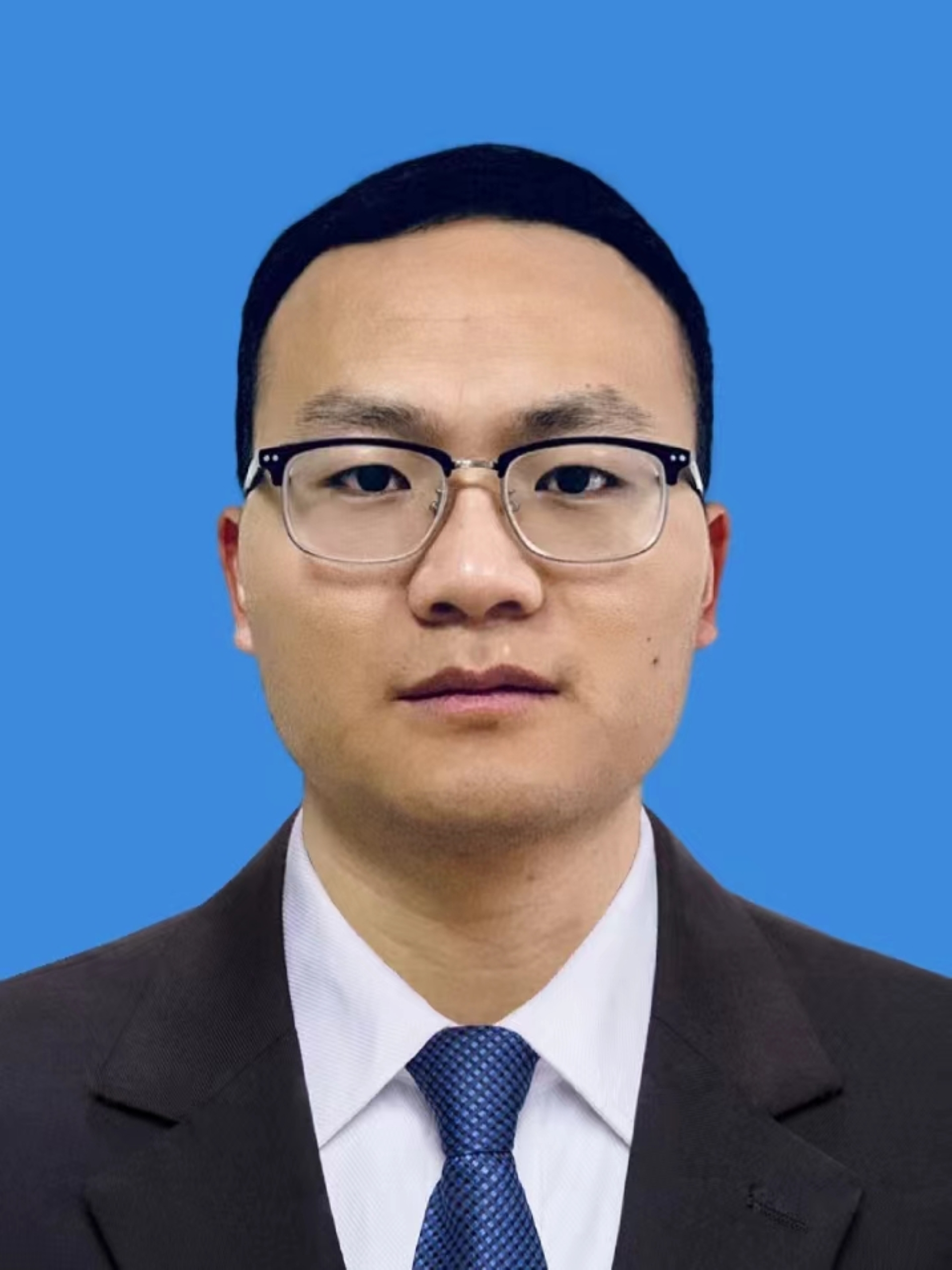}}]{Gan Sun}
(Student Member, IEEE; Member, IEEE) is currently a Full Professor with South China University of Technology. He received the B.S. degree from Shandong Agricultural University in 2013, and the Ph.D. degree from the Shenyang Institute of Automation, Chinese Academy of Sciences in 2020. He was a Visiting Student with Northeastern University from April 2018 to May 2019, and the Massachusetts Institute of Technology from June 2019 to November 2019. He was an Associate Professor with the State Key Laboratory of Robotics, Shenyang Institute of Automation, Chinese Academy of Sciences, until 2024. He has authored papers in top-tier conferences such as CVPR, ICCV, ECCV, AAAI, IJCAI, and ICDM, as well as top-tier journals including TPAMI, TNNLS, TIP, TMM, TCSVT, and Pattern Recognition. His current research interests include lifelong machine learning, multitask learning, medical data analysis, domain adaptation, deep learning, and 3D computer vision.
\end{IEEEbiography}

\begin{IEEEbiography}[{\includegraphics[width=1in,height=1.25in,clip,keepaspectratio]{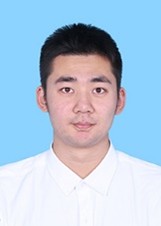}}]{Fazeng Li}
received the B.S. degree in Automation from South China University of Technology, China, in 2022, and the M.S. degree in Instrument Science and Technology from Wuhan University of Science and Technology, China, in 2025. He is currently pursuing the Ph.D. degree with South China University of Technology. His research interests include robotics, computer vision, and deep learning.
\end{IEEEbiography}

\begin{IEEEbiography}[{\includegraphics[width=1in,height=1.25in,clip,keepaspectratio]{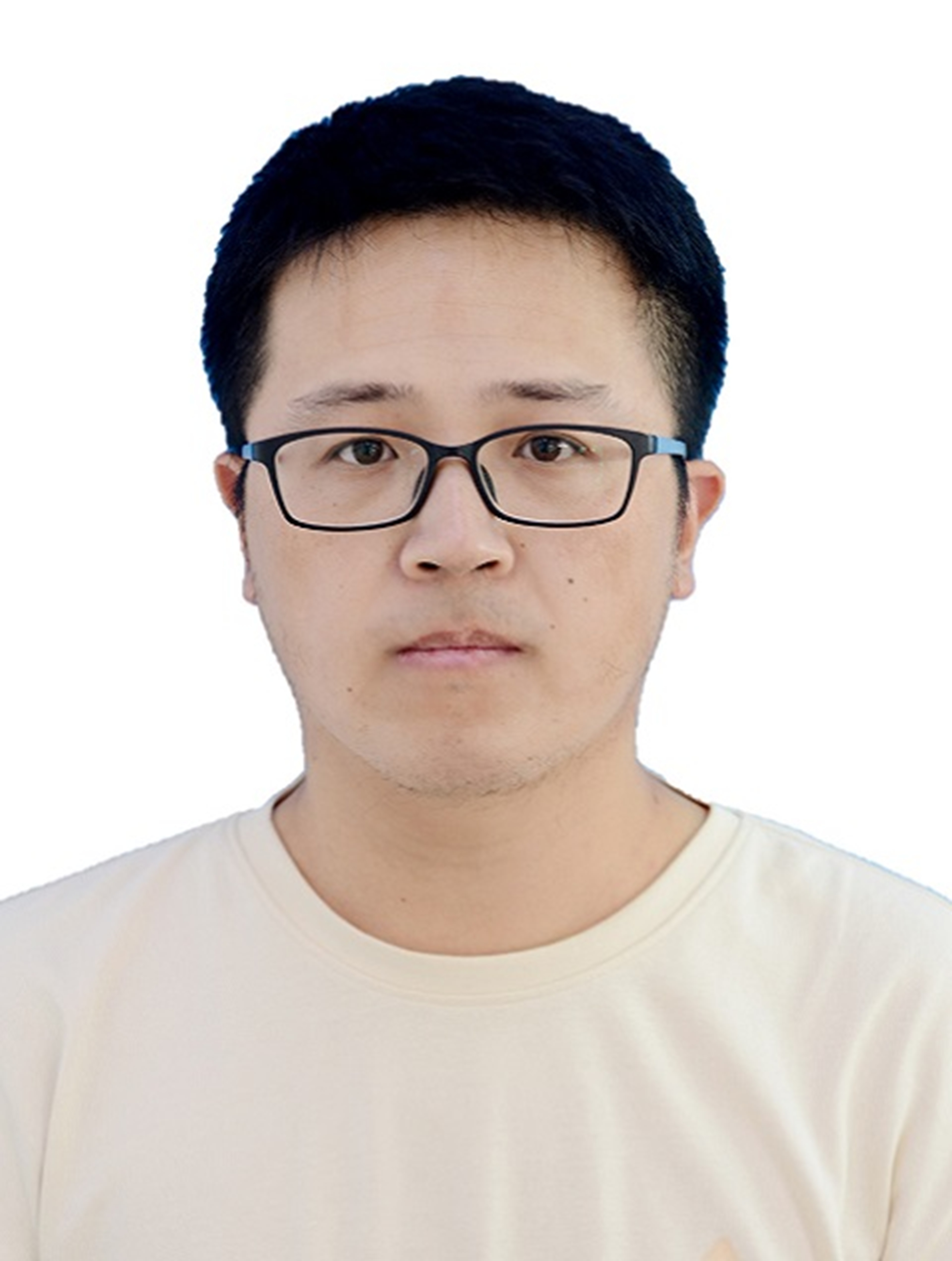}}]{Xu Tang}
received the M.E. degree in Mechanical Engineering and Automation from the Harbin Institute of Technology, Harbin, China, in 2017. He is currently pursuing the Ph.D. degree in Software Engineering with the Dalian University of Technology, Dalian, China. His current research interests include computer vision and machine learning.
\end{IEEEbiography}

\begin{IEEEbiography}[{\includegraphics[width=1in,height=1.25in,clip,keepaspectratio]{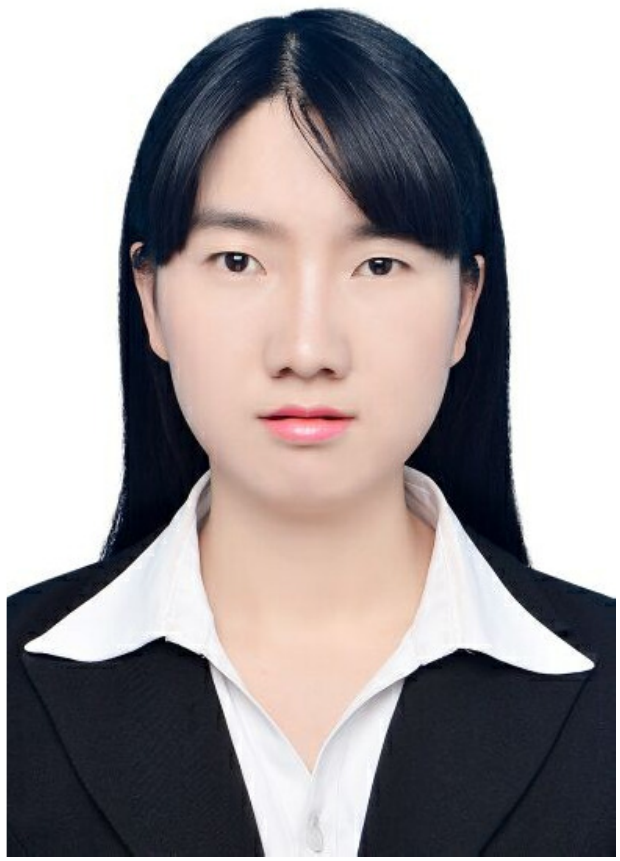}}]{Qianqian Wang}
(Member, IEEE) received the B.Eng. degree from the Lanzhou University of Technology, Lanzhou, China, in 2014, and the Ph.D. degree in pattern recognition and intelligent systems from Xidian University, Xi'an, China, in 2019. She is currently an Associate Professor with the School of Telecommunications Engineering, Xidian University. She has authored or coauthored papers in top-tier journals and conferences such as the IEEE TPAMI, IEEE TNNLS, CVPR, and AAAI. Her research interests include dimensionality reduction, pattern recognition and deep learning.
\end{IEEEbiography}

\begin{IEEEbiography}[{\includegraphics[width=1in,height=1.25in,clip,keepaspectratio]{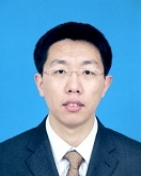}}]{Yang Cong}
(Senior Member, IEEE) received the B.Sc. degree from Northeast University in 2004, and the Ph.D. degree from the State Key Laboratory of Robotics, Chinese Academy of Sciences in 2009. He is currently a Full Professor with the School of Automation Science and Engineering, South China University of Technology, Guangzhou, China. He was a Research Fellow with the National University of Singapore (NUS) and Nanyang Technological University (NTU) from 2009 to 2011, and a Visiting Scholar with the University of Rochester. He has authored or coauthored over 100 technical papers in prestigious international journals and conferences, including the IEEE TPAMI, IEEE TNNLS, CVPR, NeurIPS, ICCV, and AAAI. His current research interests include computer vision, machine learning, lifelong learning, and medical image analysis. He serves on the editorial board of several journals.
\end{IEEEbiography}

\end{document}